\documentclass[runningheads]{llncs}
\PassOptionsToPackage{table}{xcolor}
\ifdefined\lxDocumentID\else
  \usepackage[mobile]{eccv}
\fi
\usepackage{siunitx}
\usepackage[table]{xcolor}
\usepackage{graphicx}
\usepackage{multirow}
\usepackage{booktabs}
\usepackage{colortbl} % for \cellcolor (safe)

\usepackage{tikz}
\usetikzlibrary{positioning,arrows.meta,calc}
\usepackage{adjustbox}
\usepackage{pifont} % for \ding checkmark/cross symbols
\usepackage[percent]{overpic}
\newcommand{\little}[1]{{\tiny #1}}
\definecolor{cmark-green}{RGB}{34,139,34}
\definecolor{xmark-red}{RGB}{190,30,45}
\newcommand{\cmark}{{\color{cmark-green}\ding{51}}}%
\newcommand{\xmark}{{\color{xmark-red}\ding{55}}}%
\usepackage{enumitem}

\def\dreambench{{\scshape DreamBench++}\xspace}
\newcommand{\MO}{\textsc{M--O}\xspace}                 % Metric-to-Oracle alignment (MTG)
\newcommand{\MOpair}{\textsc{M--O\textsubscript{pair}}\xspace}
\newcommand{\MH}{\textsc{M--H}\xspace}                 % Metric-to-Human alignment (DreamBench++)

\definecolor{linecolor1}{RGB}{198,239,206}  % top-1
\definecolor{linecolor2}{RGB}{236,248,239}  % top-2 (very light)
\definecolor{linecolor3}{RGB}{250,253,250}  % top-3 (almost invisible)

\newcommand{\TopOne}[1]{\cellcolor{linecolor1}{#1}}
\newcommand{\TopTwo}[1]{\cellcolor{linecolor2}{#1}}
\newcommand{\TopThree}[1]{\cellcolor{linecolor3}{#1}}

\newcommand{\twoline}[2]{\shortstack{#1\\#2}}
\newcommand{\dpos}[1]{{\scriptsize\textcolor{green!45!black}{#1}}}

\renewcommand{\twoline}[2]{\shortstack{\strut #1\\[-1.0ex] #2}}
\renewcommand{\dpos}[1]{{\tiny\textcolor{green!45!black}{#1}}}

\newcommand{\posdelta}[1]{{\scriptsize\textcolor{green!45!black}{#1}}}
\newcommand{\negdelta}[1]{{\scriptsize\textcolor{red!55!black}{#1}}}

\newcommand{\pmfmt}[2]{#1{\fontsize{6}{6}\selectfont\textcolor{black!45}{\,\,$\pm$\,\,#2}}}

\newcommand{\pmfmtB}[2]{\textbf{#1}{\fontsize{6}{6}\selectfont\textcolor{black!45}{\,\,$\pm$\,\,#2}}}

\newcommand{\tabcite}[1]{\raisebox{0pt}[1.1ex][0.2ex]{\cite{#1}}}

\makeatletter
\newcommand{\mainref}[2]{%
  \@ifundefined{r@#1}{#2}{\Cref{#1}}%
}
\makeatother

\definecolor{pretty-blue}{RGB}{0, 113, 188}
\definecolor{pretty-green}{HTML}{932232}
\definecolor{gold}{RGB}{255, 215, 0}
\definecolor{silver}{RGB}{192, 192, 192}
\definecolor{bronze}{RGB}{205, 127, 50}
\definecolor{green}{RGB}{0, 128, 0}
\definecolor{red}{RGB}{128, 0, 0}
\definecolor{blue}{RGB}{0, 0, 255}
\definecolor{linecolor1}{RGB}{211, 222, 190}
\definecolor{linecolor2}{RGB}{230, 234, 217}
\definecolor{linecolor3}{RGB}{246, 248, 239}

\definecolor{mydarkblue}{rgb}{0,0.08,0.55}
\definecolor{drp-blue}{HTML}{1f77b4}
\definecolor{mygray}{gray}{.9}
\definecolor{light-gray}{gray}{0.5}
\definecolor{kaiming-green}{RGB}{57,181,74} % kaiming green
\definecolor{icmlblue}{rgb}{0,0.08,0.45} % ICML Blue

\usepackage{eccvabbrv}
\usepackage{wrapfig}
\usepackage[accsupp]{axessibility}
\usepackage{hyperref}

\usepackage{subcaption}
\usepackage[capitalize,noabbrev]{cleveref}

\usepackage{orcidlink}

\begin{document}
% metadata.tex — Centralized title, authors, affiliations.
% \input{metadata} from main.tex, arxiv_main.tex, supp_main.tex.

\title{NearID: Identity Representation Learning via Near-identity Distractors}
\titlerunning{NearID: Identity Representation Learning via Near-identity Distractors}

\author{
Aleksandar Cvejic\inst{1}\orcidlink{0009-0005-4414-4457} \and
Rameen Abdal\inst{2}\thanks{Served in an advisory role.}\orcidlink{0000-0002-2177-8327} \and
Abdelrahman Eldesokey\inst{1}\orcidlink{0000-0003-3292-7153} \and
Bernard Ghanem\inst{1}\orcidlink{0000-0002-5534-587X} \and
Peter Wonka\inst{1}\orcidlink{0000-0003-0627-9746}
}
\authorrunning{A.~Cvejic~\orcidlink{0009-0005-4414-4457}, R.~Abdal~\orcidlink{0000-0002-2177-8327} et al.}
\institute{
King Abdullah University of Science and Technology (KAUST), Saudi Arabia \and
Snap Research, Palo Alto, CA, USA
}

\maketitle

\begin{figure}[h]
\vspace{-1cm}
    \centering
    \includegraphics[width=1.0\textwidth]{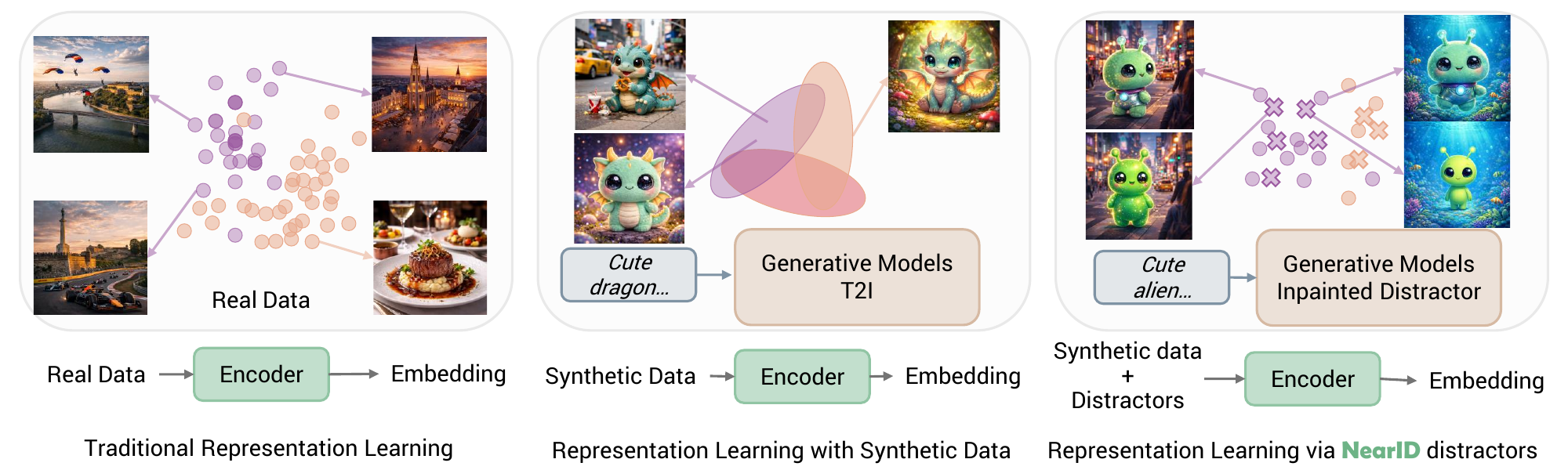}

\caption{\textbf{NearID: From Context to Identity.}
\textbf{(Left)} Traditional representations entangle object identity with background context. \textbf{(Middle)} Synthetic Data lacks explicit control over visually similar distractors.
\textbf{(Right)} \textit{NearID} introduces matched-context distractors to remove contextual shortcuts and isolate intrinsic identity signals.}
    \label{fig:teaser}
\end{figure}

% Note: doesn't change anything space wise
% but looks better with less space from teaser
\vspace{-1cm}

\begin{center}
\vspace{0.5em}
\noindent\textbf{Project page:} \url{https://gorluxor.github.io/NearID/}    
\end{center}

% \vspace{-0.5em}

% \vspace{-1cm}
\begin{abstract}

When evaluating identity-focused tasks such as personalized generation and image editing, existing vision encoders entangle object identity with background context, leading to unreliable representations and metrics. 
We introduce a principled framework for matched-context identity confusion using Near-identity (NearID) distractors, where semantically similar but distinct instances are placed on the exact same background as a reference image, eliminating contextual shortcuts and isolating identity as the sole discriminative signal. 
Based on this principle, we present the NearID dataset (19K identities, 316K matched-context distractors) together with a strict margin-based evaluation protocol. 
Under this setting, pre-trained encoders perform poorly, achieving Sample Success Rates (SSR), a strict margin-based identity discrimination metric, as low as 30.7\% and often ranking distractors above true cross-view matches. 
We address this by learning identity-aware representations on a frozen backbone using a two-tier contrastive objective enforcing the hierarchy: same identity $>$ NearID distractor $>$ random negative. 
This improves SSR to 99.2\%, enhances part-level discrimination by 28.0\%, and yields stronger alignment with human judgments on DreamBench++, a human-aligned benchmark for personalization.
% Project page: https://gorluxor.github.io/NearID/
\end{abstract}

\section{Introduction}
\label{sec:intro}

How reliably can a vision encoder tell whether two images depict the \emph{same} object? 
When the backgrounds differ, modern foundation models, including CLIP~\cite{radford2021learning_clip}, DINOv2~\cite{oquab2023dinov2}, SigLIP2~\cite{tschannen2025siglip2}, and even large vision-language models such as Qwen3-VL~\cite{bai2025qwen3}, often succeed. 
Yet these same models are systematically fooled by a simple adversarial test: replace the object with a NearID (\emph{different but visually similar}) instance while keeping the background \emph{identical}. 
Under this condition, the shared context dominates the embedding, causing the impostor to score \emph{higher} than the true identity viewed against a different backdrop. 
Such background-entangled metrics actively inflate the CLIP-I and DINO scores that the majority of personalization and editing literature relies on for automated identity preservation evaluation~\cite{peng2024dreambench++, ruiz2023dreambooth, hochberg2024towards}.

We provide a principled framework for matched-context identity confusion through Near-identity (NearID)  (See Figure~\ref{fig:teaser}): semantically similar but distinct instances inpainted into the exact same background as the reference, removing all contextual shortcuts and isolating identity as the sole discriminative signal. 
Under the resulting NearID evaluation protocol, the frozen SigLIP2 backbone achieves only $30.74\%$ Sample Success Rate (SSR), threshold-based similarity based on Near-identity distractors and positives defined in \cref{sec:NearID_protocol}, and on part-level manipulations from the Mind the Glitch (MTG) dataset~\cite{eldesokeymind_vsm_mtg}, \emph{every} standard encoder (CLIP, DINOv2, SigLIP2) scores exactly $0.0\%$ SSR.

To close this gap we introduce \textbf{NearID}, comprising three tightly coupled contributions: 
(i)~a large-scale dataset of 19k object identities with multi-view positives and over 316k NearID distractors synthesized from four generative models, providing the training signal and evaluation benchmark; 
(ii)~a two-tier contrastive objective that enforces an explicit similarity hierarchy (same identity $>$ NearID distractor $>$ random batch negative), teaching the model \emph{what} to be invariant to and \emph{what} to discriminate; 
and (iii)~a lightweight training recipe that keeps the foundation encoder frozen and learns only a Multi-head Attention Pooling (MAP)~\cite{zhai2022scaling_MAP,jaegle2021perceiver_MAP2} projection head (${\sim}3.6\%$ of total parameters), preserving the general-purpose priors while reshaping the similarity geometry for identity.

Empirically, NearID resolves the identified failure at the object level on NearID, where SSR rises from $30.74\%$ to $99.17\%$, and at the part level on Mind-the-Glitch (MTG), where SSR improves from $0.0\%$ to $35.0\%$~\cite{eldesokeymind_vsm_mtg}, a setting in which even the dedicated Visual Semantic Matching (VSM) metric reaches only $7.0\%$ under the same training data and pairwise evaluation protocol~\cite{eldesokeymind_vsm_mtg}.
Beyond SSR, we demonstrate substantially improved alignment with benchmark oracles and human judgments despite training exclusively on near-identity distractors: 
on MTG, Pearson correlation to the metric oracle increases from $0.180$ to $0.465$ under the full evaluation with view and background changes, and from $0.366$ to $0.486$ under the paired background context protocol (\cref{sec:NearID_protocol});
on \dreambench~\cite{peng2024dreambench++}, our identity-specific tuning improves overall alignment with human concept-preservation judgments even though our training data excludes style prompts and live subjects, and it generalizes to animals and humans with correlation gains of $0.105$ and $0.065$, respectively. 
These results indicate that the learned representations capture genuine identity cues rather than overfitting to synthetic artifacts.

In summary, our contributions are:
\begin{enumerate}
    \item We formalize \textbf{Near-identity distractors}: confounders that share the exact background as the reference while changing the identity signal. We introduce \textbf{NearID-bench}, a matched-context evaluation protocol (SSR and PA) that quantifies identity context entanglement in modern vision encoders, and we release a large-scale \textbf{NearID} dataset (19k identities, 316k+ distractors, 4 generative models) to support research in this setting.
    
    \item Building on the standard \textbf{MAP projection head} used with SigLIP2, we train a lightweight identity adapter on a frozen SigLIP2 backbone using a \textbf{two-tier contrastive objective} that enforces a strict similarity ordering: same identity $>$ NearID distractor $>$ random batch negative.
    
    \item We show that this training yields embeddings that substantially improve matched-context NearID distractor rejection (NearID SSR and PA), increase sensitivity to localized identity edits in part-level evaluation (MTG), and increase metric-to-human agreement on \dreambench, indicating that reducing identity context coupling leads to more reliable personalization evaluation.
\end{enumerate}

\begin{figure}[t]
    \centering
    % Replace with your actual path/filename (PDF preferred for final submission).
    % If using PNG, ensure it is high-resolution (>=300 dpi at final print size).
    \includegraphics[width=\linewidth]{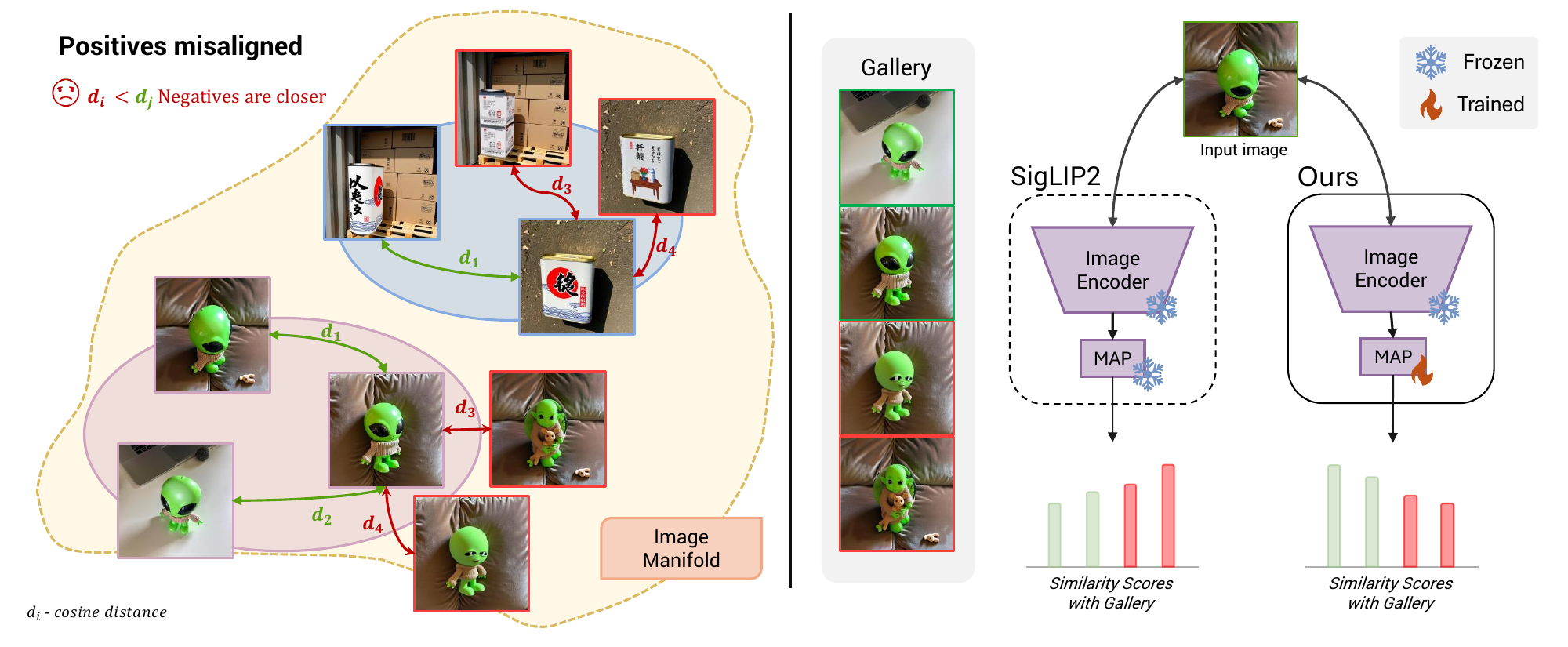}
    \caption{
    \textbf{NearID overview.}
    \textbf{Left:} In the pretrained image-embedding manifold (e.g., SigLIP2~\cite{tschannen2025siglip2}), identity-consistent \emph{positives} can be misaligned: edited or re-rendered views of the same instance may lie farther from the anchor than visually confusable \emph{negatives} (NearID distractors illustrated by $d_i < d_j$), which degrades retrieval and scoring reliability.
    \textbf{Right:} We keep the SigLIP2 image encoder frozen and train only a lightweight MAP~\cite{zhai2022scaling_MAP,jaegle2021perceiver_MAP2} projection head to reshape the similarity geometry for the target task.
    Compared to the frozen baseline, the trained head increases similarity for true positives (green) while pushing NearID distractors and unrelated gallery items lower (red), improving the desired ordering of similarity scores with respect to a gallery.
    }
    \label{fig:teaser_map_headonly}
    % \vspace{-0.5em} % (optional) tune for camera-ready if needed
\end{figure}

% =========================
% Table 1: Main table with top-1/2/3 highlights + Ours-vs-SigLIP2 deltas
% =========================
\begin{table}[t]
\centering

\caption{
\textbf{Near-identity discrimination and alignment.}
We evaluate matched-context Near-identity discrimination at the object level (NearID) and part level (MTG~\cite{eldesokeymind_vsm_mtg}), and measure correlation with MTG oracle scores and human concept-preservation judgments on DreamBench++ (DB++)~\cite{peng2024dreambench++}, verifying alignment with both oracle-defined edit severity and human perception.
\textbf{Top:} Existing embedding and VLM methods under the Near-identity protocol.
\textbf{Bottom:} Training in our setting, showing improvements on NearID and gains on MTG and DB++.
MTG reports metric-to-oracle alignment (\MO/\MOpair); DB++ reports metric-to-human alignment (\MH).
Deltas are relative to the frozen SigLIP2 baseline.
* denotes a method trained directly on MTG.
SSR and PA are averaged across seven inpainting settings (three excluded from training; see Sec.~\ref{sec:NearID_protocol}).
}

\label{tab:main_table}
\renewcommand{\arraystretch}{1.15}
\setlength{\tabcolsep}{2.5pt}
\scriptsize
\begin{tabular}{lccccccc}
\toprule
\multirow{2}{*}{\textbf{Scoring Model}} &
\multicolumn{2}{c}{\textbf{NearID}} &
\multicolumn{4}{c}{\textbf{MTG}} &
\multicolumn{1}{c}{\textbf{DB++}} \\
\cmidrule(lr){2-3}\cmidrule(lr){4-7}\cmidrule(lr){8-8}
& \textbf{SSR} $\uparrow$ & \textbf{PA} $\uparrow$ &
\textbf{\MO} $\uparrow$ & \textbf{\MOpair} $\uparrow$ &
\textbf{SSR} $\uparrow$ & \textbf{PA} $\uparrow$ &
\textbf{\MH} $\uparrow$ \\
\midrule
Qwen3VL30B~\cite{bai2025qwen3}&
\TopTwo{49.73} & \TopTwo{69.20} &

0.219 & \TopTwo{0.329} &
\TopThree{17.0} & \TopThree{26.0} &
-- \\
CLIP~\cite{radford2021learning_clip}         &
10.31 & 20.92 &
0.239 & 0.484 &
0.0 & 0.0 &
\TopThree{0.493} \\
DINOv2~\cite{oquab2023dinov2}       &
20.43 & 34.55 &
\TopTwo{0.324} & \TopOne{\textbf{0.519}} &
0.0 & 0.0 &
0.492 \\
VSM*~\cite{eldesokeymind_vsm_mtg}          &
\TopThree{32.13} & 46.70 &
\TopTwo{0.394} & 0.445 &

\TopTwo{7.0} & \TopTwo{24.5} &
0.190 \\ % \textit{[fill]}
\midrule
SigLIP2~\cite{tschannen2025siglip2}      &
30.74 & \TopThree{48.81} &
0.180 & 0.366 &
0.0 & 0.0 &
\TopTwo{0.516} \\
% \addlinespace[1.5pt]
\textbf{NearID (Ours)} &
\TopOne{\twoline{\textbf{99.17}}{\dpos{(+68.43)}}} &
\TopOne{\twoline{\textbf{99.71}}{\dpos{(+50.90)}}} &
\TopOne{\twoline{\textbf{0.465}}{\dpos{(+0.285)}}} &
\TopThree{\twoline{0.486}{\dpos{(+0.120)}}} &
\TopOne{\twoline{\textbf{35.0}}{\dpos{(+35.0)}}} &
\TopOne{\twoline{\textbf{46.5}}{\dpos{(+46.5)}}} &
\TopOne{\twoline{\textbf{0.545}}{\dpos{(+0.029)}}} \\
\bottomrule
\end{tabular}
\end{table}

\section{Related Work}
\label{sec:related_work}
\noindent\textbf{Visual foundation models and identity representation:} 
Representation learning has advanced rapidly through self-supervised~\cite{caron2021emerging_dino, he2022masked_MAE, grill2020bootstrap_BYOL, chen2020simple_SimCLR, fang2023eva}, multi-modal~\cite{radford2021learning_clip, li2022blip, jia2021scaling_ALIGN, Yu2022CoCaCC, alayrac2022flamingo}, and generative pretraining~\cite{rombach2022high_sd, ruiz2023dreambooth, wang2024instantid, wang2025editcliprepresentationlearningimage}, driving progress across discriminative~\cite{li2022Grounded_GLIP, kirillov2023segment} and generative tasks. 
Modern vision-language models (VLMs)~\cite{liu2023visualinstructiontuning, li2023blip2, dai2023instructblip, wang2024cogvlmvisualexpertpretrained, bai2025qwen3, achiam2023gpt} are now widely adopted as general-purpose encoders~\cite{zhai2023SigLIP, tschannen2025siglip2, simeoni2025dinov3, woo2023ConvNeXtV2}. Contrastive language-image models~\cite{radford2021learning_clip,tschannen2025siglip2} excel at broad category-level alignment, and the self-supervised DINO family~\cite{oquab2023dinov2,caron2021emerging_dino} learns powerful dense features and patch-level structural representations. 
However, because these models are trained for broad semantic grouping, they often entangle instance identity with context~\cite{singhania2025BeyondPixels}, yielding high similarity for images with similar backgrounds despite different instances. Recent bias analyses confirm that VLM behavior can be strongly influenced by background patterns, and that removing the background can substantially reduce such biases~\cite{vo2025vision}. 
Specialized identity encoders~\cite{deng2019arcface, papantoniou2024arc2face, Kim2022AdaFaceQA, meng2021magfaceuniversalrepresentationface, schroff2015facenet} achieve strong identity separation but remain confined to narrow domains such as human faces. 
Region-focused approaches like AlphaCLIP~\cite{sun2023alphaclipclipmodelfocusing} augment CLIP with an auxiliary alpha channel to extract features from user-specified regions (e.g., masks or bounding boxes) while retaining contextual awareness, but still lack explicit identity-level supervision. 
In contrast, our work seeks to extract open-domain, instance-level identity representations. 
Rather than risking catastrophic forgetting by fine-tuning a foundation model, we leverage the robust priors of a frozen backbone and utilize a lightweight Multi-head Attention Pooling (MAP) head~\cite{zhai2022scaling_MAP,jaegle2021perceiver_MAP2}, which we train to explicitly attend to identity-salient features while suppressing contextual background cues.

\vspace{1mm}\noindent\textbf{Deep metric learning and hierarchical contrastive objectives:} 
Standard deep metric learning relies on pulling positive pairs together while pushing negatives apart. Traditional formulations utilize batch-based contrastive losses~\cite{oord2018representation_CPC,chen2020simple_SimCLR} or supervised extensions that group multiple positives SupContrast~\cite{khosla2020supervised_SupContrast}. 
While effective for instance discrimination, these binary objectives treat all negatives equally, forcing the model to push semantically related intra-class (NearID) samples and completely unrelated random samples to the same near-zero similarity. 
To inject structure into the negative space, prior works have explored triplet mining~\cite{schroff2015facenet}, multi-negative objectives~\cite{sohn2016improved}, and dynamic pair weighting~\cite{wang2019multi, sun2020circle_loss}. 
Hierarchical metric learning further attempts to impose discrete geometric boundaries using ranked lists or tree-based proxies~\cite{wang2019multi_MS_loss,yang2022hierarchical_loss}. 
However, these methods typically rely on online algorithmic selection of informative negatives, which is unstable during training, or on predefined categorical taxonomies. Our approach differs fundamentally by utilizing \emph{explicitly curated} near-identity distractors generated via controlled inpainting. 
We enforce a strict three-tier structural hierarchy (positives $>$ Near-identity negatives $>$ batch negatives) through our dual-tier objective, providing stable, constant gradient pressure without the instability of algorithmic mining.
Closest to our use of contrastive signals for identity, CustomContrast~\cite{chen2025customcontrast} employs a multi-level contrastive objective for subject-driven generation, but co-trains an encoder \emph{within} a text-to-image generator; in contrast, NearID learns a standalone, generator-agnostic identity \emph{metric} purely from near-identity distractor margins.
\noindent\textbf{Continuous calibration and learning-to-rank:} 
Beyond strict categorical ranking, identity preservation often exhibits varying degrees of severity, requiring continuous calibration of similarity scores. 
Soft contrastive learning and contrastive regression frameworks~\cite{thoma2020soft_learning,keramati2023conr,zha2023rank_n_contrast} replace binary targets with continuous variables, adjusting the contrastive push based on label distance. 
Similarly, Learning-to-Rank (LTR) objectives optimize for monotonic sorting across retrieved items rather than absolute distances~\cite{burges2005learning_to_rank,cao2007learning,xia2008listwise}. 
While post-hoc calibration strategies like temperature scaling or isotonic regression~\cite{guo2017calibration,zadrozny2002transforming} can adjust score distributions after training, they cannot fix underlying manifold collisions. 
In contrast to methods requiring explicit continuous regression targets, our approach achieves data-driven structural calibration. 
By subjecting a diverse curriculum of part-level edits (via the MTG dataset~\cite{eldesokeymind_vsm_mtg}) to our ranking regularizer ($\mathcal{L}_{\text{rank}}$), the embedding magnitudes naturally correlate with physical edit severity. 
This smoothly interpolates between identical instances and unrelated batch negatives without relying on manual numeric margin during optimization.

\noindent\textbf{Evaluation of personalized image generation:} 
The rapid advancement of subject-driven generation and image editing~\cite{ruiz2023dreambooth, gal2022textual_inversion, ye2023ip_adapter, wang2024instantid, qian2025omni-id, cvejic2025partedit, mao2025realcustom++, wu2025less} has exposed a critical vulnerability in current evaluation protocols. 
The vast majority of personalization literature relies on out-of-the-box CLIP~\cite{radford2021learning_clip} cosine similarity (CLIP-I) or DINO-scores to quantify identity preservation~\cite{hochberg2024towards}.
However, as our experiments demonstrate, these generic semantic metrics are easily fooled by near-identity confounders, leading to inflated scores when a model generates a different instance on the correct background. 
Recent work by Kilrain et al.~\cite{kilrain2025finer} further shows that pairwise CLIP/DINO similarities can remain near-perfect when layout and background match while identity-defining details drift, and proposes a gallery-based retrieval protocol to expose this failure mode. 
In personalization pipelines, this identity and context coupling manifests in both learning and inference: optimizing textual identifiers with standard diffusion losses can entangle foreground and background information~\cite{Chen2026ConceptCraftOP}, and diffusion-transformer architectures with global attention can conflate semantic identity with spatial layout~\cite{hu2025positionic}. 
Current evaluation protocols using VLMs as judges further compound these issues, since they provide the full image as input rather than isolating the subject~\cite{ku2024viescore,zhang2023gpt4visiongeneralistevaluatorvisionlanguage}.

Recent efforts to bridge this gap have introduced perceptual metrics calibrated to human judgments, ranging from low-level distortion penalties~\cite{zhang2018lpips} to mid-level holistic similarity metrics like DreamSim~\cite{fu2023dreamsim}. 
While DreamSim successfully aligns with human preferences regarding layout, pose, and overall composition, it evaluates \textit{holistic} scene similarity rather than isolating \textit{strict object identity}. Consequently, it remains susceptible to background conflation when a structurally similar distractor shares the reference context. 
Furthermore, recent analysis by PercepAlign~\cite{sundaram2024does} demonstrates that fully fine-tuning vision encoders to align with generic human perceptual judgments actively degrades their underlying high-level semantic representations. 

Our proposed framework addresses these dual challenges. 
To prevent representation degradation, we freeze the foundation backbone and tune only a lightweight MAP head to project features into a specialized identity subspace. 
Concurrently, our NearID protocol provides the first rigorous, automated evaluation standard explicitly designed to penalize background conflation and reward true object-level identity preservation rather than holistic visual similarity.
% =========================
% Combined figure: Attention maps (left) + Comparison table (right)
% Left: Placeholder for attention map visualization (Deliverable 3)
% Right: Property comparison table (comparison.tex)
% =========================

% preamble:
% \usepackage{subcaption}

% preamble:
% \usepackage{subcaption}
% \usepackage{graphicx}   % for \rotatebox
% \usepackage{booktabs}   % for \toprule etc.

% ---------- Figure ----------
\begin{figure*}[t]
  \centering

  \begin{subfigure}[t]{0.38\textwidth}
    \centering
    \vspace{0pt}
    \includegraphics[width=0.9\linewidth]{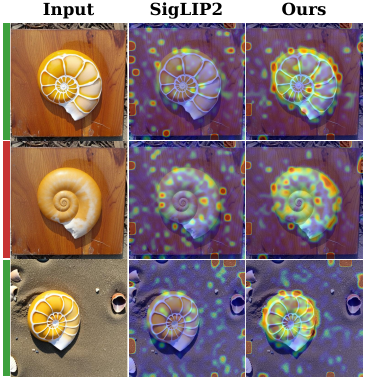}
  \end{subfigure}\hfill
  \begin{subfigure}[t]{0.60\textwidth}
    \centering
    \vspace{0.5cm}

    \scriptsize
    \renewcommand{\arraystretch}{1.15}
    \setlength{\tabcolsep}{4pt}

    \begin{tabular}{@{}l c c c c@{}}
      \toprule
      % headroom for rotated labels (no dummy row)
      \rule{0pt}{3.0ex} &
      \rotatebox{65}{\textbf{DINOv2}} &
      \rotatebox{65}{\textbf{SigLIP2}} &
      \rotatebox{65}{\textbf{VSM}} &
      \rotatebox{65}{\textbf{NearID}} \\
      \midrule
      NearID object-level & \xmark & \xmark & \xmark & \cmark \\
      NearID part-level   & \xmark & \xmark & \cmark & \cmark \\
      Mask-free            & \cmark & \cmark & \xmark & \cmark \\
      Human alignment      & \cmark & \cmark & \xmark & \cmark \\
      Oracle alignment     & \xmark & \xmark & \cmark & \cmark \\
      \bottomrule
    \end{tabular}
  \end{subfigure}

  \caption{\textbf{Left:} Attention map comparison vs baseline on positives and negatives. \textbf{Right:} Summary of properties evaluated in this paper. NearID improves matched-context Near-ID rejection on NearID-bench and transfers to part-level identity evaluation on MTG, while remaining mask-free at inference.}
  \label{fig:attn_and_comparison}
  \vspace{-0.5cm}
\end{figure*}

\section{NearID}
\label{sec:nearid}

\subsection{Overview and Problem Formulation}
\label{sec:nearid_overview}

We formulate identity preservation as a structured metric learning problem designed to explicitly disentangle object identity from background context (see Figure~\ref{fig:teaser_map_headonly}). \textbf{NearID} is a unified framework comprising three tightly coupled components: 
(i) a hierarchical contrastive objective that enforces an explicit similarity ordering between positives, near-identity distractors, and random negatives (\Cref{sec:problem_formulation}), 
(ii) a large-scale matched-context dataset with curated identities and synthesized near-identity distractors that instantiate this training setting (\Cref{sec:dataset_synthesis}), and 
(iii) a rigorous evaluation protocol based on discriminability margins and alignment with human and oracle judgments (\Cref{sec:NearID_protocol}). 

Together, these elements isolate identity as the sole discriminative signal and enable principled training and reliable evaluation under matched-context confounders.

\paragraph{Training Tuple Construction.}
Each training sample is constructed as a tuple $\mathcal{T}_i = \{ a_i, \{g_{i,p}\}_{p=1}^{P}, \{r_{i,k}\}_{k=1}^{K} \}$
where $a_i$ denotes an anchor image of object identity $i$, $g_{i,p}$ are up to $P$ positive views of the \emph{same} identity against different backgrounds, $r_{i,k}$ are $K$ \emph{near-identity distractors}. A NearID distractor consists of a semantically similar but distinct object instance that is inpainted into the \emph{exact same background} as the anchor $a_i$. This matched-context construction removes all contextual shortcuts and forces the model to discriminate based solely on intrinsic identity cues.

\paragraph{Learning Objective.}
The goal is to learn an embedding function $f_\theta : \mathcal{I} \rightarrow \mathbb{R}^d$ such that, for a given anchor $a_i$, in terms of cosine similarity,
$\cos(a_i, p_i)\;>\;\cos(a_i, n_i^{\text{near}})\;>\;\cos(a_i, n_i^{\text{rand}})$,
i.e., $\mathbf{\text{same identity}} \;>\; \mathbf{\text{NearID distractor}} \;>\; \mathbf{\text{random batch negative}}$.
This explicitly enforces a structured similarity hierarchy rather than treating all negatives equally.

\paragraph{Parameter-Efficient Adaptation.}
To avoid catastrophic forgetting and preserve the robust semantic priors of large-scale vision encoders, we freeze a pre-trained backbone (SigLIP2~\cite{tschannen2025siglip2}) and optimize only a lightweight Multi-head Attention Pooling (MAP) projection head~\cite{zhai2022scaling_MAP, jaegle2021perceiver_MAP2}. 

Given spatial patch embeddings extracted from the frozen backbone, the MAP head selectively aggregates identity-salient features and projects them into a refined metric space. This design updates only $\sim$3.6\% of total parameters while reshaping the embedding geometry to satisfy the identity hierarchy defined above (See Figure~\ref{fig:attn_and_comparison} left pane). In the following subsection, we formalize the extended hierarchical contrastive objective used to enforce this structure.

\subsection{NearID Loss}
\label{sec:problem_formulation}

Standard contrastive objectives treat all negatives uniformly, failing to penalize background-induced correlations that allow models to shortcut on contextual cues rather than intrinsic identity features (illustrated in Figure~\ref{fig:teaser} and Figure~\ref{fig:teaser_map_headonly}).
We address this with the \textbf{NearID loss} ($\mathcal{L}_{\text{NearID}}$), a two-component objective that enforces a structured three-level similarity hierarchy: true identity matches above NearID distractors, and NearID distractors above generic batch negatives.

\paragraph{Notation.}
Let $\phi(x)$ denote frozen backbone features and let $f$ denote the trainable MAP projection head applied on top of $\phi$.
For an image $x$, we define the $\ell_2$-normalized embedding $\mathbf{z}_x = \frac{f(\phi(x))}{\|f(\phi(x))\|_2} \in \mathbb{R}^d$.
All pairwise comparisons use temperature-scaled similarity logits $\ell_{u,v} = \frac{\mathbf{u}^\top \mathbf{v}}{\tau}$, where $\tau > 0$ is a fixed temperature hyperparameter ($\tau = 0.07$).
Since embeddings are $\ell_2$-normalized, dot products equal cosine similarities.

For each anchor image $a_i$, $\{g_{i,p}\}_{p=1}^{P}$ positive views, and $\{r_{i,k}\}_{k=1}^{K}$ NearID distractors (See Section~\ref{sec:nearid_overview}), we define their embeddings as $\mathbf{a}_i = \mathbf{z}_{a_i}$, $\mathbf{g}_{i,p} = \mathbf{z}_{g_{i,p}}$, and $\mathbf{r}_{i,k} = \mathbf{z}_{r_{i,k}}$. Let $\mathcal{G} = \{\mathbf{g}_{j,q}\}$ denote the global set of all positive embeddings gathered across devices (DDP), with $|\mathcal{G}| = M$.
Define the anchor's own positive set $\mathcal{G}_i = \{\mathbf{g}_{i,q}\}_{q=1}^{P} \subset \mathcal{G}$ and the batch-negative pool $\mathcal{B}_{\text{neg}}^{(i)} = \mathcal{G} \setminus \mathcal{G}_i$. Note that the NearID distractor set for anchor $i$, 
$\mathcal{R}_i = \{\mathbf{r}_{i,k}\}_{k=1}^{K}$, 
is defined per-anchor and is not included in $\mathcal{G}$.
In practice, we average only over valid positives and distractors using masks (omitted for clarity).

\paragraph{Discrimination term ($\mathcal{L}_{\text{disc}}$).}
For each positive index $p$, the discrimination term applies a softmax cross-entropy over the global positive pool augmented with the NearID distractors as  {\small $\mathcal{L}_{\text{disc},p}^{(i)} = -\log \frac{\exp(\ell_{\mathbf{a}_i, \mathbf{g}_{i,p}})}{\sum_{g \in \mathcal{G}} \exp(\ell_{\mathbf{a}_i, g}) + \sum_{k=1}^{K} \exp(\ell_{\mathbf{a}_i, \mathbf{r}_{i,k}})}.$}

This formulation is structurally related to InfoNCE~\cite{oord2018representation_infonce_loss} and multi-class N-pair objectives~\cite{sohn2016improved}, but with two key differences.
First, the NearID distractors $\mathbf{r}_{i,k}$ are explicitly included in the denominator alongside global batch candidates, providing targeted identity-confusing competition that standard batch negatives cannot supply.
Second, unlike Supervised Contrastive Learning~\cite{khosla2020supervised_SupContrast}, which places all same-class positives in the numerator, we retain the other $P{-}1$ positive views in the denominator.
This per-index formulation mitigates multi-positive collapse and ensures that each view is individually discriminable under background variation.
The discrimination loss averages over valid positive indices:

\begin{equation}
\small
\mathcal{L}_{\text{disc}} \;=\; \frac{1}{P} \sum_{p=1}^{P} \mathbb{E}_i \!\left[\mathcal{L}_{\text{disc},p}^{(i)}\right].
\label{eq:disc_avg}
\end{equation}

\paragraph{Ranking regularizer ($\mathcal{L}_{\text{rank}}$).}
The discrimination term alone treats NearID distractors simply as additional negatives to push away.
Under strong gradient pressure from highly confusable distractors, this can collapse the local semantic neighborhood, destroying the graded similarity structure that distinguishes ``nearby but different'' from ``unrelated'', a known concern in metric learning with highly confusable negatives~\cite{wang2019multi_MS_loss}. To preserve this structure, we add a ranking regularizer that encourages each NearID distractor to be ranked \emph{above} the generic batch-negative pool.
Define the batch-negative log-sum-exp as $\text{LSE}_i \;=\; \log \sum_{g \in \mathcal{B}_{\text{neg}}^{(i)}} \exp(\ell_{\mathbf{a}_i, g}).$ which acts as a smooth maximum over the batch-negative pool.

For each NearID distractor $r_{i,k}$, the ranking regularizer takes the form, $\mathcal{L}_{\text{rank}}^{(i,k)} \;=\; \log\!\Big(1 + \exp\!\big(\text{LSE}_i - \ell_{\mathbf{a}_i, \mathbf{r}_{i,k}}\big)\Big).$

which is a softplus penalty that decreases smoothly as the NearID distractor logit exceeds the aggregate batch-negative signal, and increases when batch negatives dominate.
This is equivalent to the cross-entropy form, $\mathcal{L}_{\text{rank}}^{(i,k)} \;=\; -\log \frac{\exp(\ell_{\mathbf{a}_i, \mathbf{r}_{i,k}})}{\exp(\ell_{\mathbf{a}_i, \mathbf{r}_{i,k}}) \;+\; \sum_{g \in \mathcal{B}_{\text{neg}}^{(i)}} \exp(\ell_{\mathbf{a}_i, g})}.$

However, the softplus view makes the listwise ranking nature explicit: $\text{LSE}_i$ acts as a smooth maximum over the batch-negative pool, and the penalty enforces that each NearID distractor logit exceeds this threshold.
Aggregating:
\begin{equation}
\small
\mathcal{L}_{\text{rank}} \;=\; \frac{1}{K} \sum_{k=1}^{K} \mathbb{E}_i \!\left[\mathcal{L}_{\text{rank}}^{(i,k)}\right].
\label{eq:rank_avg}
\end{equation}

\paragraph{NearID loss.}
The complete objective combines discrimination and ranking:
\begin{equation}
\small
\mathcal{L}_{\text{NearID}} \;=\; \mathcal{L}_{\text{disc}} \;+\; \alpha\, \mathcal{L}_{\text{rank}},
\label{eq:nearid_loss}
\end{equation}
where $\alpha = 0.5$ controls the ranking regularizer strength.
Together, the two terms enforce the intended similarity ordering $\ell_{\mathbf{a}_i, \mathbf{g}_{i,p}} \;>\; \ell_{\mathbf{a}_i, \mathbf{r}_{i,k}} \;\gtrsim\; \text{LSE}_i.$

ensuring that true positives are selected over all candidates ($\mathcal{L}_{\text{disc}}$), while NearID distractors remain closer than generic batch negatives ($\mathcal{L}_{\text{rank}}$), preserving the graded semantic structure of the embedding space.

\subsection{NearID Dataset Construction}
\label{sec:dataset_synthesis}

% =======================================================================
% Figure: Contrastive Overview — SimCLR vs StableRep vs NearID
% Promoted to main paper (Sec. 3.3). Short caption here; the full
% per-paradigm discussion lives in supp.tex (Sec. A).
% =======================================================================
\begin{figure}[t]
    \centering
    \includegraphics[width=\linewidth]{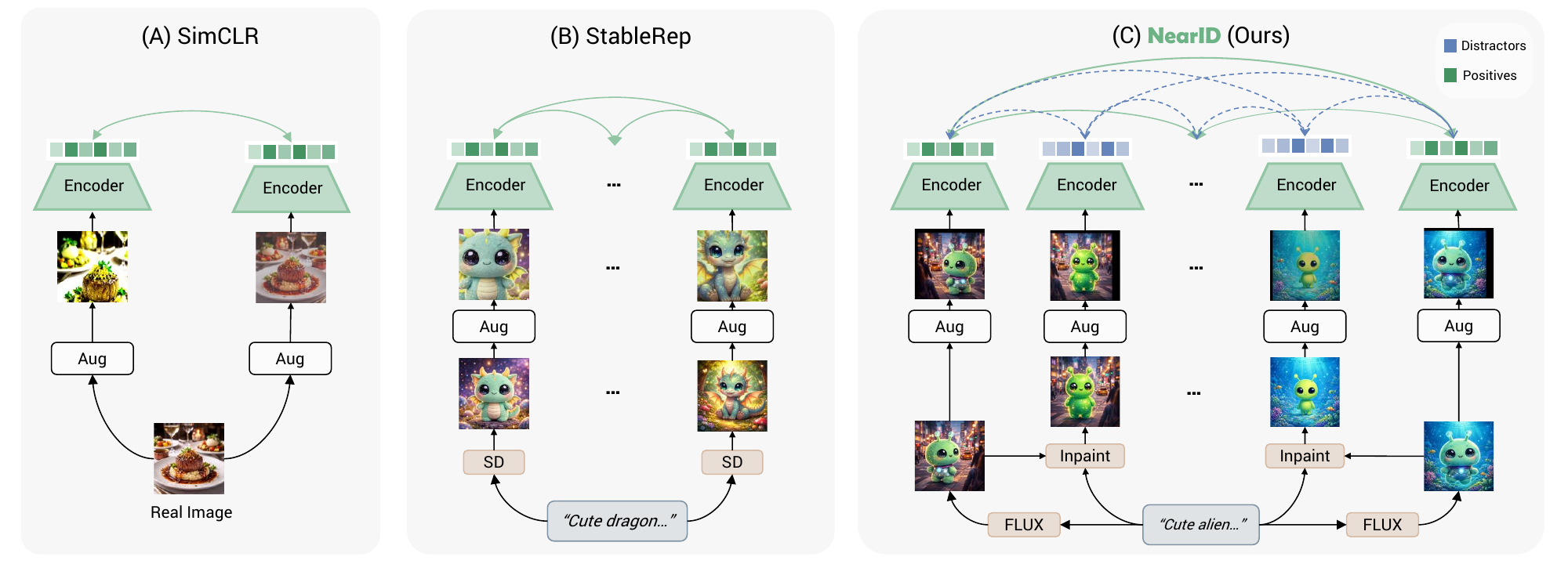}
    \caption{
    \textbf{Contrastive training paradigms.}
    Unlike \textbf{(A)}~SimCLR~\cite{chen2020simple_SimCLR} (augmented views of a single image) and \textbf{(B)}~StableRep~\cite{tian2023stablerep} (same-caption generations), \textbf{(C)}~NearID uses multi-view positives of the \emph{same} instance and adds \textit{near-identity distractors}, visually similar but distinct instances inpainted into the anchor's background, a structured negative signal absent from both. See the supplement for a full discussion.
    }
    \label{fig:contrastive_overview}
\end{figure}

\vspace{1mm}\noindent\textbf{Source datasets.}
\textbf{SynCD}~\cite{kumari2025generating_syncd} renders up to three views of the \emph{same} instance across backgrounds from Objaverse~\cite{deitke2022objaverseuniverseannotated3d} 3D assets via conditioned diffusion; all such views, including front and back, are positives, as identity rather than viewpoint defines a positive pair. We extend SynCD by adding \textit{near-identity distractors} as a structured negative signal absent from prior contrastive paradigms (\cref{fig:contrastive_overview}). \textbf{MTG}~\cite{eldesokeymind_vsm_mtg} pairs the \emph{same} object under a localized part edit, providing a fine-grained calibration signal.

\vspace{1mm}\noindent\textbf{Dataset Foundation and Curation.}
Constructing a robust benchmark for object-level identity requires data with high viewpoint diversity and strict identity separation. 
While datasets such as Mind-The-Glitch (MTG)~\cite{eldesokeymind_vsm_mtg} and Subjects200k~\cite{tan2025ominicontrol_subjects200k} offer valuable baselines, they are limited by scale (e.g., 5K samples in MTG), a lack of multi-view diversity, and a focus on subject-level matching against a single plain-background reference, rather than instance-level object identity under viewpoint and background changes.
Furthermore, MTG explicitly cites significant dataset quality issues within Subjects200k, which further motivates our decision to improve upon it. 

To address these limitations, we build our synthesis pipeline on the SynCD dataset \cite{kumari2025generating_syncd} (90K initial samples).
To guarantee absolute identity fidelity and prevent semantic overlap during training and evaluation, we apply a rigorous filtering protocol. 
We first discard the ``deformable'' partition (approximately half of the dataset) because the publicly released version lacks the exact foreground text descriptions required for precise generative conditioning and low diversity with 16 base classes. 
Within the remaining ``rigid'' object subset, we enforce strict uniqueness constraints based on paired textual descriptions and \texttt{ObjaverseID}~\cite{deitke2022objaverseuniverseannotated3d}. 
This curation process yields our proposed dataset, \textbf{NearID}, comprising 19,386 unique object identities and 45,215 distinct images. Notably, 12,943 identities (67\%) possess two distinct views, and 6,433 identities (33\%) provide three distinct views. 
This scale and viewpoint coverage significantly surpasses the entire MTG dataset (limited to 5K identities with only 2 views). 
We create mutually exclusive train/val/test splits with 18,786/100/500 identities, respectively.

\vspace{1mm}\noindent\textbf{Multi-Model Synthesis Pipeline.}
Using this curated core subset, we generate near-identity distractors employing a diverse ensemble of SoTA diffusion models. 
This pipeline encompasses Stable Diffusion XL (SDXL) \cite{podell2023sdxl}, FLUX.1 \cite{flux2024}, Qwen-Image \cite{wu2025qwenimage}, and PowerPaint \cite{zhuang2024task_powerpaint} (utilizing the BrushNet v2.1 pipeline \cite{ju2024brushnet}), which ensures a wide variance in generative priors, structural adherence, and artifact distributions. 
Overall, this yields $7$ inpainting configurations and 316,505 near-identity distractor images, disentangling inpainting artifacts, model-specific biases, and source-dependent generative fingerprints.

To supplement our object-level distractors with fine-grained variations, we concatenate the MTG dataset, which inherently consists of part-level inpaintings from a single source (SDXL). 
To ensure equal sampling probability during training, the MTG subset is repeated four times to match the scale of our synthesized data. 
Generation is conducted using mixed precision, where supported by the model architecture, to optimize throughput during large-scale dataset construction. 
We use default parameters from publicly available code for each method, with slight modifications when utilizing adapters. Exact inference hyperparameters are detailed in the supplementary material.

\vspace{1mm}\noindent\textbf{Data Processing Pipeline.} 
Generation is performed at two distinct resolutions: $512^2$ and the respective model-native resolutions (e.g., spatial scaling to align the shortest side with $640$ for SD 1.5-based variants like BrushNet, or strictly $1024^2$ for SDXL), except for BrushNet, which operates exclusively at its native resolution. 
Following the synthesis of the near-identity distractors, spatial standardization is enforced by downsampling any high-resolution outputs to $512^2$ utilizing Lanczos resampling. This ensures consistent dimensions for downstream evaluation metrics and contrastive training while mitigating aliasing artifacts.

\subsection{Evaluation Protocol}
\label{sec:NearID_protocol}

Given the tuple formulation introduced in \Cref{sec:nearid_overview}, we evaluate identity discrimination under a matched-context setting designed to eliminate background shortcuts. The protocol analyzes similarity margins within strict identity–background triplets. For a given identity, let $\mathbf{p}_i$ and $\mathbf{p}_j$ denote two positive views of the same object captured against distinct backgrounds $i$ and $j$. Let $\mathbf{n}_i$ denote a NearID distractor: a semantically similar but distinct instance inpainted into the exact background of $\mathbf{p}_i$.
\paragraph{Bidirectional Discriminability Margin.}
We measure whether identity similarity across backgrounds exceeds similarity to a matched-context distractor. For a pair $(i,j)$, we define the directed margins:
\begin{equation}
    \delta^{(ij)}_{i \rightarrow j} = s(\mathbf{p}_i, \mathbf{p}_j) - s(\mathbf{p}_i, \mathbf{n}_i),
    \qquad
    \delta^{(ij)}_{j \rightarrow i} = s(\mathbf{p}_i, \mathbf{p}_j) - s(\mathbf{p}_j, \mathbf{n}_j).
\end{equation}
A margin trial is considered successful if $\delta > 0$, indicating that cross-background identity similarity overrides matched-context confounders.

\paragraph{Micro-pooled Metrics.}
We aggregate directed margin trials using two measures:

\begin{description}[leftmargin=0pt, font=\bfseries]
    \item[Sample Success Rate (SSR).]
    A sample is counted as successful only if \emph{all} valid directed margins are positive (up to six per identity):
    \item[Pairwise Accuracy (PA).]
    The proportion of successful directed margin trials across the dataset, treating each margin independently:
\end{description}
\begin{equation}
\small
\text{SSR} = \frac{1}{N}\sum_{i=1}^{N} \mathbb{1}\!\Big[\,\forall\,(j,d):\; \delta_{d}^{(ij)} > 0\,\Big], \quad
\text{PA} = \frac{\big|\{m \in \mathcal{M} : \delta_m > 0\}\big|}{|\mathcal{M}|},
\label{eq:ssr_pa}
\end{equation}
where $N$ is the number of identities and $\mathcal{M}$ is the set of all directed margin trials.

\paragraph{Alignment with Human and Oracle Judgments.} To ensure our learned embeddings align with fine-grained identity preservation, we evaluate against the MTG dataset \cite{eldesokeymind_vsm_mtg} and DreamBench++ (DB++) \cite{peng2024dreambench++}. 
We utilize the MTG part-level oracle score, defined as $1 - (|\Omega_{p}| / |\Omega_{o}|)$, where $\Omega_{p}$ and $\Omega_{o}$ represent the part and object masks, respectively. Hence, we also compute two Pearson~\cite{pearson1895vii} correlations:

\begin{description}[leftmargin=0pt, font=\bfseries]
    \item[\textbf{M -- O}.] 
    Correlations between the cosine similarities and the oracle scores.
    \item[\textbf{M -- H}.] 
    Correlations between the cosine similarities and human scores on DB++.
\end{description}

All correlations are reported as the mean Pearson correlation, $\bar{r}$, computed via Fisher's $z$-transformation~\cite{fisher1915frequency_zfisher} over per-sample coefficients (see supplementary material for the aggregation formula).

\begin{figure*}[t]
    \centering
    \includegraphics[width=\linewidth]{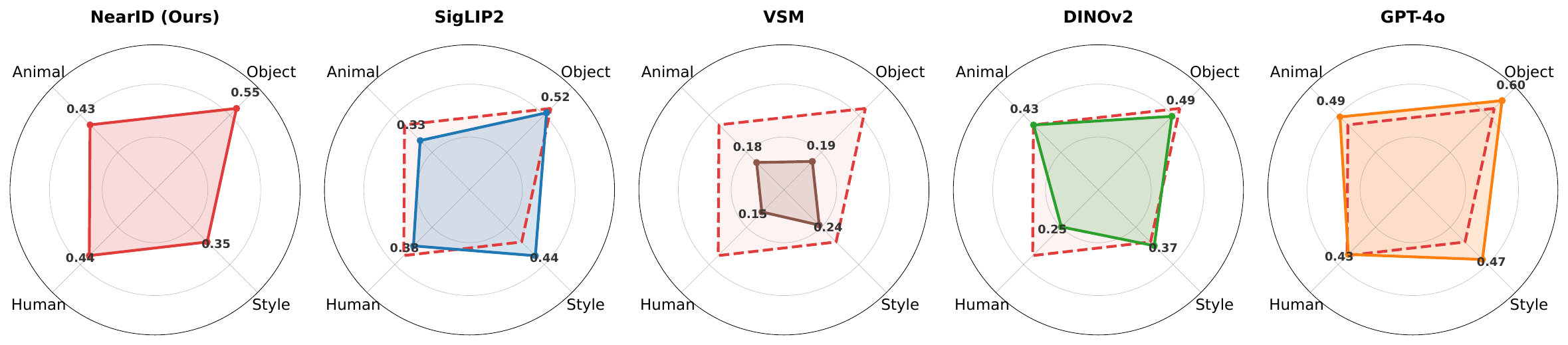}
    \caption{
    \textbf{Per-category human alignment ($M-H$) on DreamBench++~\cite{peng2024dreambench++}.}
    Each radar shows the Pearson correlation between a metric and human concept-preservation judgments across four DB++ categories; \textcolor[HTML]{E03C3C}{NearID (Ours)} is repeated as a dashed reference in each subplot.
    Despite training exclusively on rigid objects, NearID improves over the frozen SigLIP2 baseline on \emph{Animal} ($+0.105$) and \emph{Human} ($+0.065$), indicating that disentangling identity from context transfers across semantic domains.
    The expected decrease on \emph{Style} ($-0.092$), a category entirely absent from training, confirms that the gains reflect genuine identity learning rather than general score inflation.
    }
    \label{fig:subcategory}
\end{figure*}

\begin{table}[!t]
\caption{
\textbf{\dreambench concept-preservation human alignment.}
Pearson correlations between human (H) and metric scores:
G = GPT, D = DINO, C = CLIP, S = SigLIP2, NearID (Ours). Bolded numbers indicate improvement over SigLIP2.
}
\label{tab:alignment_cp}
\resizebox{\linewidth}{!}{
\begin{tabular}{lcccccccc}
\toprule
\multirow{2}{*}[-0.5ex]{\textbf{Method}} &
\multirow{2}{*}[-0.5ex]{\textbf{T2I Model}} &
\multicolumn{6}{c}{\textbf{Concept Preservation} (Metric--Human Alignment)} \\
\cmidrule(lr){3-8}
& & \textbf{H--H} & \textbf{G--H} & \textbf{D--H} & \textbf{C--H} & \textbf{S--H} & \textbf{NearID--H} \\
\midrule

\textbf{Textual Inversion}~\tabcite{gal2022textual_inversion} & SD v1.5
& 0.684 & \pmfmt{0.625}{0.032} & \pmfmt{0.611}{0.016} & \pmfmt{0.610}{0.001} & \pmfmt{0.660}{0.003} & \pmfmtB{0.675}{0.007} \\
\textbf{DreamBooth}~\tabcite{ruiz2023dreambooth} & SD v1.5
& 0.731 & \pmfmt{0.722}{0.011} & \pmfmt{0.632}{0.010} & \pmfmt{0.623}{0.015} & \pmfmt{0.652}{0.012} & \pmfmtB{0.655}{0.023} \\
\textbf{DreamBooth LoRA}~\tabcite{ruiz2023dreambooth,hu2022lora} & SDXL v1.0
& 0.651 & \pmfmt{0.637}{0.004} & \pmfmt{0.446}{0.013} & \pmfmt{0.481}{0.007} & \pmfmt{0.480}{0.006} & \pmfmtB{0.546}{0.006} \\
BLIP-Diffusion~\tabcite{li2023blip} & SD v1.5
& 0.630 & \pmfmt{0.466}{0.019} & \pmfmt{0.311}{0.024} & \pmfmt{0.290}{0.004} & \pmfmt{0.363}{0.016} & \pmfmtB{0.382}{0.002} \\
Emu2~\tabcite{sun2024generative_emu2} & SDXL v1.0
& 0.763 & \pmfmt{0.777}{0.004} & \pmfmt{0.774}{0.018} & \pmfmt{0.744}{0.021} & \pmfmt{0.766}{0.021} & \pmfmtB{0.768}{0.028} \\
IP-Adapter-Plus ViT-H~\tabcite{ye2023ip_adapter} & SDXL v1.0
& 0.530 & \pmfmt{0.424}{0.045} & \pmfmt{0.212}{0.017} & \pmfmt{0.262}{0.002} & \pmfmt{0.253}{0.002} & \pmfmtB{0.306}{0.018} \\
IP-Adapter ViT-G~\tabcite{ye2023ip_adapter} & SDXL v1.0
& 0.485 & \pmfmt{0.401}{0.021} & \pmfmt{0.261}{0.004} & \pmfmt{0.283}{0.031} & \pmfmt{0.248}{0.019} & \pmfmtB{0.316}{0.041} \\

\midrule
\textbf{Fisher-$z$ ($\bar{r}$ - \cref{sec:eval})} & &
\pmfmt{0.648}{0.038} &
\TopOne{\pmfmt{0.597}{0.056}} &
\pmfmt{0.492}{0.081} &
\pmfmt{0.493}{0.074} &
\TopThree{\pmfmt{0.516}{0.079}} &
\TopTwo{\pmfmtB{0.545}{0.071}} \\
\bottomrule
\end{tabular}
}
\vspace{-10pt}
\end{table}

\section{Experiments}
\label{sec:experiments}

\subsection{Implementation Details}
\label{sec:impl_details}

\paragraph{Architecture and Optimization.} 
All embedding models in our proposed pipeline share a frozen \texttt{SigLIP2-so400m-patch14-384} backbone, unless otherwise specified. 
We train only the Multihead Attention Pooling (MAP) projection head, which outputs a $1152$ dimensional $\ell_2$-normalized embedding. 
This head-only tuning regime updates approximately 15M parameters out of the roughly 428M total model parameters. 
By training merely ${\sim}3.6\%$ of the overall network capacity, we ensure minimal computational overhead while preserving the robust zero-shot priors of the foundation model. 
Models are optimized using AdamW ($\eta=10^{-4}$, weight decay=$10^{-4}$) in mixed-precision (\texttt{fp16}). 
We employ a cosine annealing schedule with a 100-step linear warmup, utilizing a global batch size of 128 across 11 epochs (approximately $3,350$ gradient steps).

\paragraph{Role-Aware Data Augmentation.}
We apply role-aware stochastic foreground masking: backgrounds are blacked out for anchors ($p{=}0.5$), positives ($p{=}0.2$), and distractors ($p{=}0.6$), the latter to prevent trivial rejection via background border artifacts.
Positives receive the lightest masking because their cross-background variation is itself the identity signal; a mask-free ablation (\mainref{sec:supp_role_masking}{see supplementary material}) confirms this masking is essential for background-invariant, part-level transfer.
Standard augmentations (color jitter, flipping, translation/scale) are applied per slot.

\paragraph{Joint Training.} The NearID dataset is interleaved with the MTG training split (upsampled $4\times$).
MTG's part-level edits are subjected to $\mathcal{L}_{\text{rank}}$ without explicit regression targets, enabling continuous calibration from the data distribution.

\begin{table}[t]
\centering
\caption{
\textbf{Training objective ablation.}
All variants share a frozen SigLIP2 backbone with a trainable MAP head, trained on NearID\,+\, MTG data for 11~epochs with identical hyperparameters.
More details in the supplementary material about the specifics of each loss and the extension.
$^\dagger$Representation collapse: high discrimination but degraded general-purpose alignment (see text). We observe that our configurations pose the best balance between the different adaptations to the NearID setting.
}
\label{tab:ablation}
\renewcommand{\arraystretch}{1.15}
\setlength{\tabcolsep}{1.5pt}
\scriptsize
\begin{tabular}{lccccccc}
\toprule
\multirow{2}{*}{\textbf{Training Loss}} &
\multicolumn{2}{c}{\textbf{NearID}} &
\multicolumn{4}{c}{\textbf{MTG}} &
\multicolumn{1}{c}{\textbf{DB++}} \\
\cmidrule(lr){2-3}\cmidrule(lr){4-7}\cmidrule(lr){8-8}
& \textbf{SSR} $\uparrow$ & \textbf{PA} $\uparrow$ &
\textbf{\MO} $\uparrow$ & \textbf{\MOpair} $\uparrow$ &
\textbf{SSR} $\uparrow$ & \textbf{PA} $\uparrow$ &
\textbf{\MH} $\uparrow$ \\
\midrule

None (frozen)
  & 30.74 & 48.81
  & 0.180 & 0.366 & 0.0 & 0.0
  & 0.516 \\
\midrule

InfoNCE \little{(sym., 1-pos)} 
  & 60.97 & 75.26
  & 0.267 & 0.418 & 8.0 & 12.0
  & 0.555 \\
~~+ $\mathcal{R}$ neg 
  & 99.57 & 99.79
  & 0.236 & 0.267 & 64.0 & 74.0
  & 0.251 \\
~~+ Oracle Ranking$^\dagger$ 
  & 86.34 & 92.25
  & 0.299$^\dagger$ & 0.444$^\dagger$ & 7.0$^\dagger$ & 11.0$^\dagger$
  & 0.167$^\dagger$ \\
~~+ $\mathcal{R}$ neg + Oracle Ranking 
  & 99.60 & 99.89
  & 0.247 & 0.277 & 65.0 & 74.5
  & 0.227 \\
\addlinespace[2pt]

SigLIP \little{(BCE, ext., + hard>batch rank)} 
  & 61.40 & 77.81
  & 0.213 & 0.385 & 6.0 & 9.5
  & 0.530 \\
Circle \little{(ext.,+hard>batch margin rank)}$^\dagger$ % R4: W&B yufe4xk6
  & \textit{99.97} & \textit{99.99}
  & 0.264$^\dagger$ & 0.303$^\dagger$ & \textit{67.0}$^\dagger$ & \textit{76.5}$^\dagger$
  & 0.141$^\dagger$ \\
\addlinespace[2pt]
% ---- (c) NearID loss (ours) ----
\textbf{$\mathcal{L}_{\textbf{NearID}}$ (Ours)} 
  & \textbf{99.17} & \textbf{99.71}
  & \textbf{0.465} & \textbf{0.486} & 35.0 & 46.5
  & \textbf{0.545} \\
~~+ Pos.\ Cohesion 
  & 99.31 & 99.78
  & 0.459 & 0.485 & 36.0 & 47.0
  & 0.541 \\
\bottomrule
\end{tabular}
\end{table}

\subsection{Results}
\label{sec:eval}

We evaluate against frozen contrastive models (CLIP~\cite{radford2021learning_clip}, SigLIP2~\cite{tschannen2025siglip2}), self-supervised transformers (DINOv2~\cite{oquab2023dinov2}), a VLM (Qwen3-VL\,30B~\cite{bai2025qwen3}), and VSM~\cite{eldesokeymind_vsm_mtg}, trained on MTG.
SSR and PA are pooled across distractor sources via support-weighted averaging.

As shown in \Cref{tab:main_table}, frozen encoders fail under matched-context evaluation: SigLIP2 achieves only 30.74\% SSR.
NearID resolves this, reaching 99.17\% SSR and 99.71\% PA.
On MTG, all standard encoders score 0.0\% SSR; even VSM reaches only 7.0\%.
NearID achieves 35.0\% SSR with substantially improved oracle alignment (\MO: 0.465 vs.\ 0.180).
On DB++ (\Cref{tab:alignment_cp}), NearID improves metric-to-human correlation to 0.545 vs.\ 0.516 for SigLIP2.
As shown in \Cref{fig:subcategory}, these gains generalize across DB++ categories, including animals and humans absent from training, confirming that identity-specific training transfers to real-world evaluation.
\Cref{fig:attn_and_comparison} right pane summarizes the resulting property profile across methods.

\begin{figure}[t]
    \centering
    \includegraphics[width=\linewidth]{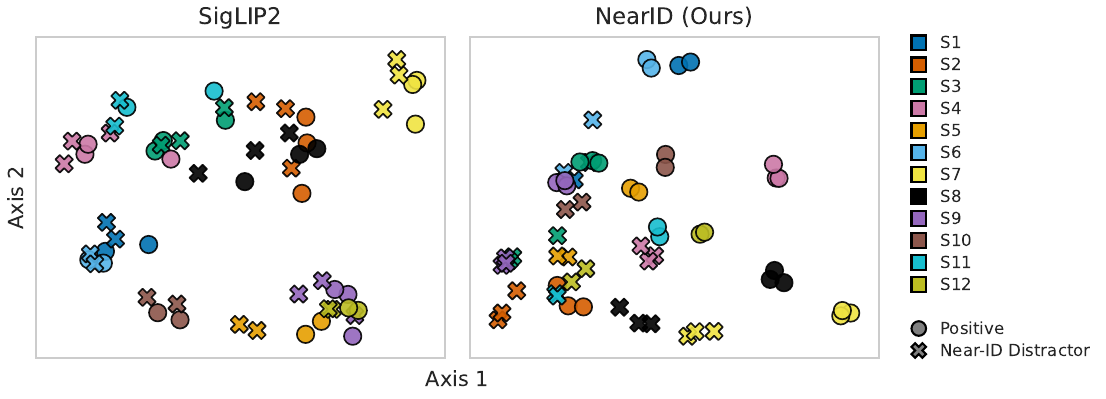}
    \caption{
    \textbf{KernelPCA visualization of Near-Identity separation.}
    We project embeddings for $n{=}12$ identities (S1--S12) into 2D using identical KernelPCA~\cite{scholkopf1998nonlinear_kernelpca} settings in both panels.
    Circles denote positives (same identity) and crosses denote matched-context NearID distractors (negatives).
    Compared to the frozen SigLIP2 baseline, NearID increases separation by pushing the distractors away from the corresponding positive clusters, providing visual evidence of improved near-identity discrimination.
    }
    \label{fig:kpca_nearid}

\end{figure}

\subsection{Visualization of Embeddings}
\label{sec:visualization_of_embeddings}

\Cref{fig:kpca_nearid} shows KernelPCA~\cite{scholkopf1998nonlinear_kernelpca} projections of NearID validation embeddings.
The frozen baseline (left) entangles distractors with positives, while NearID (right) separates them into distinct clusters, with positives grouped but individually discriminable.

\subsection{Training Objective Ablation}
\label{sec:ablation}

We train the same MAP head on identical data with different loss functions (\Cref{tab:ablation}); only the loss differs.

\paragraph{Standard contrastive training is insufficient.}
InfoNCE improves SSR from 30.74\% to 60.97\% and achieves the highest \MH\,(0.555), but 61\% SSR means near-identity distractors still frequently outscore true positives.

\paragraph{Hierarchy is necessary but collapse is a risk.}
Circle\,+\,Ranking achieves 99.97\% NearID SSR but \textit{collapses} to \MH\,=\,0.141, well below the 0.516 frozen baseline.
Oracle Ranking similarly collapses (\MH\,=\,0.167).
These aggressive formulations over-specialize the embedding, motivating the softer softplus formulation in $\mathcal{L}_{\text{rank}}$.

\paragraph{NearID loss balances discrimination and alignment.}
$\mathcal{L}_{\text{NearID}}$ attains 99.17\% SSR while maintaining \MH\,=\,0.545, trading only 0.010 \MH relative to InfoNCE (which fails at 61\% SSR).

Adding positive cohesion yields marginal gains (99.31\% SSR) with comparable MTG SSR ($36.0\%$ vs $35.0\%$), confirming that the base $\mathcal{L}_{\text{NearID}}$ suffices for the primary setting.

\section{Conclusion}
\label{sec:conclusion}

We presented \textbf{NearID}, a unified framework for learning identity-discriminative representations under matched-context confounders.
We showed that widely used vision encoders systematically conflate object identity with background context: under our evaluation protocol, even the best frozen encoder achieves only 30.74\% SSR when near-identity distractors share the reference background.

To address this, we introduced three tightly coupled components: (i)~a large-scale dataset with 19K identities and 316K+ matched-context distractors from four generative models, (ii)~a two-component contrastive objective that enforces a strict similarity hierarchy without hard margins or oracle supervision, and (iii)~a discriminability-based evaluation protocol grounded in similarity margins.
Training only a lightweight MAP head on a frozen backbone, NearID achieves 99.17\% SSR, improves part-level discrimination on MTG, and increases alignment with human judgments on DreamBench++.

The NearID dataset and evaluation protocol can serve as a diagnostic tool for any vision encoder, providing a rigorous test of whether learned representations capture genuine identity or merely exploit contextual shortcuts.

\vspace{1mm}\noindent\textbf{Limitations.}
Our scope targets common rigid objects rather than specialized domains such as faces or industrial inspection; distractor realism is bounded by the fidelity of the inpainting models used to synthesize them; and NearID is never trained on stylized data, leaving style-conditioned identity to future work.

We hope NearID establishes a stronger standard for evaluating and learning identity-aware representations in personalized generation and image editing.

\section*{Acknowledgements}
The research reported in this publication was supported by funding from King Abdullah University of Science and Technology (KAUST) - Center of Excellence for Generative AI, under award number 5940 and Snap Research. Rapidata ai for their API access.
%and a gift from Google.

\ifdefined\lxDocumentID
  \bibliographystyle{unsrt}% splncs04 is unknown to LaTeXML
\else
  \bibliographystyle{splncs04}
\fi
\bibliography{main}

% supp
\clearpage
% =======================================================================
% supp.tex --- Supplement body (no preamble, no \begin{document})
%
% This file is \input{} by:
%   - supp_main.tex  (standalone supplementary PDF for ECCV submission)
%   - arxiv_main.tex (combined paper + supplement for arXiv)
%
% Do NOT add \documentclass, package loading, or \begin{document} here.
% =======================================================================

% Fix duplicate hyperref destinations when main + supp are combined (arxiv_main.tex)
\renewcommand{\theHsection}{supp.\Alph{section}}
\appendix

\section{Contrastive Training Paradigm Comparison}
\label{sec:supp_contrastive_overview}

NearID sits within the broader landscape of contrastive representation learning (\mainref{fig:contrastive_overview}{see the main-paper figure}).
SimCLR~\cite{chen2020simple_SimCLR} constructs positive pairs by applying strong stochastic augmentations (random cropping, color distortion, Gaussian blur) to a single real image, encouraging pixel-level invariance via pairwise InfoNCE.
StableRep~\cite{tian2023stablerep} replaces hand-crafted augmentations with generative diversity: a text-to-image diffusion model produces multiple images from the same caption under different noise seeds, and these caption-conditioned samples form multi-positive sets optimised with a supervised contrastive~\cite{khosla2020supervised_SupContrast} objective, yielding caption-level invariance.

NearID differs from both paradigms along two axes.
\textbf{(i)}~Positives are neither augmented copies of a single image nor caption-conditioned generations; they are high-quality multi-view images produced by the SynCD~\cite{kumari2025generating_syncd} pipeline, which leverages Objaverse~\cite{deitke2022objaverseuniverseannotated3d} 3D assets as a geometric prior: depth-conditioned diffusion generation (e.g., FLUX.1-Depth) combined with cross-view feature warping ensures strict 3D-consistent multi-view positives (up to three views per identity; \mainref{sec:dataset_synthesis}{Section~3.3}).
\textbf{(ii)}~NearID introduces \textit{near-identity distractors}, visually similar yet non-matching instances inpainted into the same background as the anchor, as a structured negative signal that is absent from both SimCLR and StableRep.
The NearID loss (\mainref{eq:nearid_loss}{Eq.\,3}) leverages these distractors to enforce a three-level similarity ordering ($\text{identity} > \text{distractor} > \text{batch negative}$), preserving the graded semantic structure rather than treating all non-matching samples with uniform repulsion. We note, however, that the naive solution of placing near-identity distractors directly in the softmax denominator, without the ranking regulariser, achieves near-perfect discrimination (\Cref{tab:infonce_ablation}) but destroys alignment with both oracle scores and human judgments in the process.

\section{Dataset Construction Details}
\label{sec:supp_dataset}

\subsection{Inpainting Details}

\vspace{1mm}\noindent\textbf{Synthesis Engines and Hyperparameters.} To maximize the photorealism and contextual coherence of the generated distractors, each model is configured to its optimal operational regime:

\begin{itemize}
    \item \textbf{PowerPaint (BrushNet Variant):} We utilize the PowerPaint-v2 architecture, built upon a Stable Diffusion v1.5 backbone (Realistic Vision V6.0) and integrated with a BrushNet conditioning model~\cite{ju2024brushnet} for disentangled mask-image guidance. Generation is driven by learned task tokens (e.g., text-guided object synthesis) to precisely govern the inpainting behavior. Inference is executed over $45$ denoising steps utilizing the UniPC Multistep Scheduler, a classifier-free guidance (CFG) scale of $7.5$, and a fitting degree (brushnet conditioning scale) of $1.0$.

    \item \textbf{SDXL-Inpainting:} We utilize the \texttt{SDXL-1.0-inpainting-0.1} checkpoint operating at $1024^2$ resolution. Distractor synthesis is performed over $30$ denoising steps with a CFG scale of $8.0$ and an inpainting strength of $0.99$ to ensure maximal adherence to the unmasked context. A comprehensive negative prompt is applied to suppress text, watermarks, and low-resolution artifacts.

    \item \textbf{FLUX.1 Frameworks:} We leverage the guidance-distilled FLUX.1-dev ecosystem to generate structurally consistent distractors. Masked region synthesis (\textbf{FLUX.1-Fill}) is executed for $50$ steps with a high guidance scale of $30.0$. For topology-preserving structural edits, we employ \textbf{FLUX.1-Canny-Inpaint} configured with $28$ steps, a guidance scale of $7.0$, and an inpainting strength of $0.99$, guided by edge maps extracted via dual thresholds ($\tau_{low}=50, \tau_{high}=200$). Negative prompting is inherently omitted for all FLUX.1 variants.

    \item \textbf{Qwen-Image (ControlNet):} We integrate the InstantX ControlNet with the \texttt{Qwen-Image} backbone for accelerated distractor generation. To facilitate high-throughput dataset creation, we fuse an 8-step Lightning LoRA (\texttt{Qwen-Image-Lightning-8steps-V1.1}) and utilize the Flow-Match Euler Discrete Scheduler, reducing the inference trajectory to $8$ steps and the true CFG scale to $1.0$.
\end{itemize}

\section{Additional Qualitative Results}
\label{sec:supp_qualitative}

\begin{figure}[t]
\centering
\includegraphics[width=\linewidth]{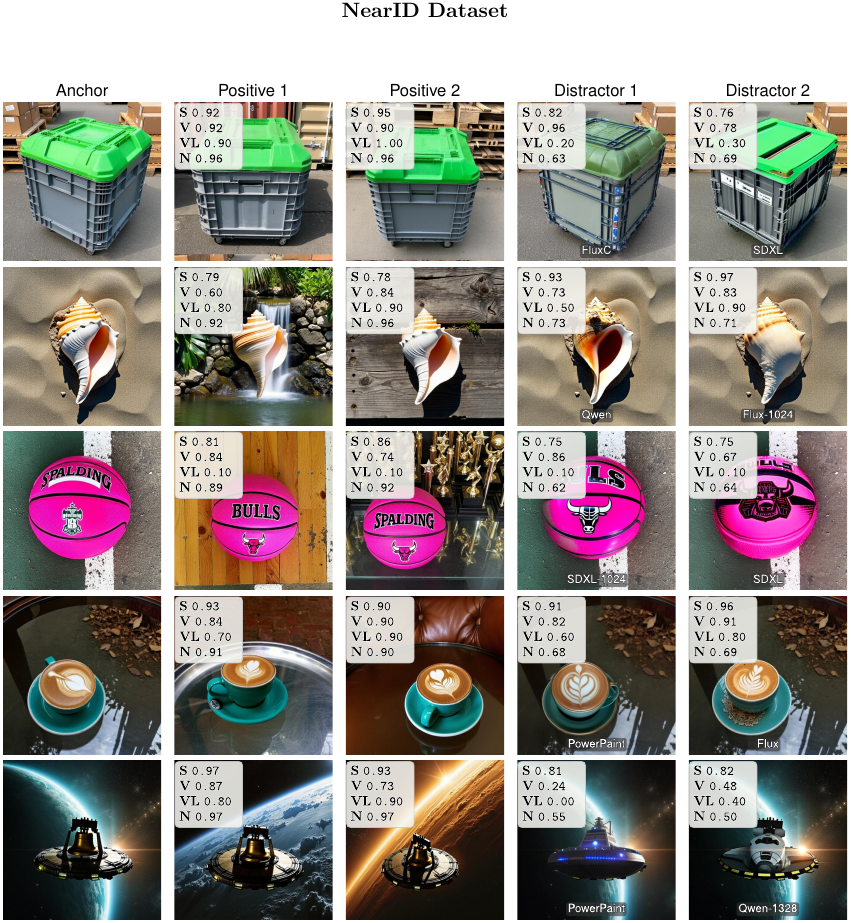}
\caption{
\textbf{NearID qualitative results.}
Each row shows one identity: Anchor\,|\,Positive\,1\,|\,Positive\,2\,|\,Distractor\,1\,|\,Distractor\,2.
Positives depict the same instance in a different background; near-identity distractors show a different but visually similar instance inpainted into the anchor's original background.
\textbf{NearID (N)} consistently assigns higher scores to positives than to distractors across diverse object categories (storage crates, conch shells, basketballs, coffee cups, spacecraft).
In contrast, frozen SigLIP2 (\textbf{S}) frequently assigns distractor scores on par with or exceeding positives; for instance, in the seashell row \textbf{S}~rates Distractor~2 at $0.97$, actually exceeding Positive~1 ($0.79$), while \textbf{N} correctly suppresses the distractor to $0.71$.
The VLM judge (\textbf{VL}) is inconsistent: near-zero scores for the basketball row ($0.10$) yet $1.00$ for a spacecraft distractor, reflecting the VLM's tendency toward category-level rather than instance-level reasoning.
NearID successfully resolves these confounds, maintaining a clear margin between positives and near-identity distractors in all five categories shown.
}
\label{fig:supp_nearid_qualitative}
\end{figure}

\begin{figure}[t]
\centering
\includegraphics[width=\linewidth]{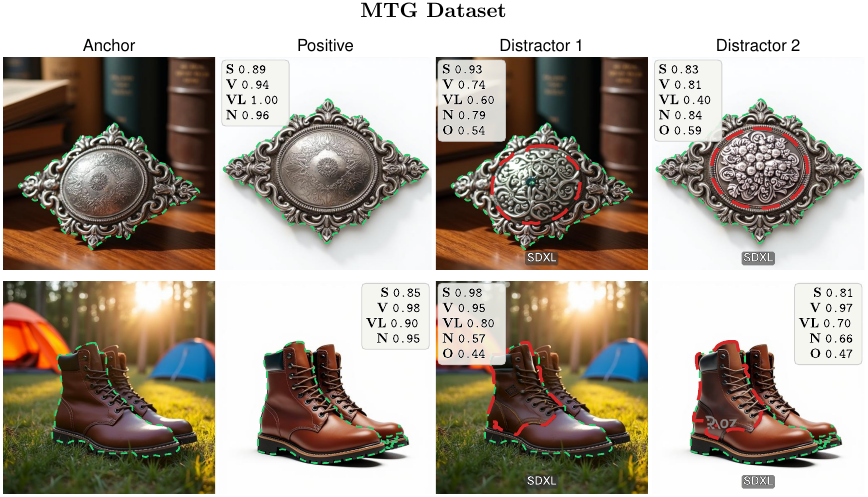}
\caption{
\textbf{MTG qualitative results.}
Each row shows one identity: Anchor\,|\,Positive\,|\,Distractor\,1\,|\,Distractor\,2.
Distractors are generated by inpainting a localized part edit (e.g., the central motif of a brooch, the toe-cap design of a boot) while preserving the overall object shape and background; Oracle (\textbf{O}) measures the fractional area of the edited region.
The red contour marks the edited part boundary; the green contour traces the full object extent.
In the ornate brooch row, frozen SigLIP2 (\textbf{S}~$= 0.93$) actually ranks Distractor~1 \emph{higher} than the genuine Positive ($0.89$), a failure caused by overall appearance similarity.
\textbf{NearID (N)} correctly ranks Positive ($0.96$) above both distractors ($0.79$, $0.84$), detecting the fine-grained motif difference despite near-identical backgrounds.
In the boot row, VSM (\textbf{V}) assigns Distractor~2 a score of $0.97$, nearly identical to the Positive ($0.98$), whereas NearID cleanly separates them ($0.95$ vs.\ $0.66$) even though the Oracle score ($0.47$) confirms a moderate but genuine edit.
These examples illustrate that NearID captures part-level identity differences beyond what holistic embedding models detect.
}
\label{fig:supp_mtg_qualitative}
\end{figure}

We present qualitative comparisons on both the NearID and MTG evaluation sets.
For each sample we show per-image similarity scores for four metrics:
\textbf{S}\,=\,SigLIP2 (frozen), \textbf{V}\,=\,VSM, \textbf{VL}\,=\,Qwen3-VL\,30B, and \textbf{N}\,=\,NearID (ours).
Scores are computed relative to the anchor; the anchor itself shows no score. We can see visualization of MTG in \cref{fig:supp_mtg_qualitative} and NearID in \cref{fig:supp_nearid_qualitative}.

% \subsection*{MTG Dataset}

\section{Computational Cost}
\label{sec:supp_compute}

Training the NearID MAP head (15M trainable parameters) for 11 epochs (${\sim}3{,}350$ gradient steps, batch size 128) requires approximately 6.5 hours on a single NVIDIA A100 (80\,GB) in mixed-precision (\texttt{fp16}), at roughly 7\,s per step including periodic checkpointing and evaluation callbacks.
Because the SigLIP2 backbone is frozen and only the lightweight MAP head is updated, the compute budget is modest: a single run costs ${\sim}6.5$ A100-hours, and the full ablation suite (9 loss variants $+$ 5 data-split runs $+$ 5 $\alpha$-sweep runs) totals ${\sim}124$ A100-hours of training.
The additional $\beta$-sweep (3 cohesion variants at $\alpha{=}1.0$) contributes a further ${\sim}20$ A100-hours.
Offline evaluation of a single NearID checkpoint across all 9 distractor sources and the MTG benchmark takes approximately 10 minutes on one A100, adding negligible overhead.

\paragraph{VLM baseline cost.}
The Qwen3-VL\,30B VLM baseline is substantially more expensive to evaluate.
Each NearID sample requires a separate autoregressive VLM inference with structured JSON output (${\sim}260$ tokens), consuming ${\sim}72$\,GB of VRAM on an A100 (each sample comparisons are batched with a fixed seed).
Evaluating a single distractor source ($500$ test samples) takes approximately 3 hours, and the full NearID benchmark (9 sources $\times$ 2 masking conditions) requires ${\sim}54$ A100-hours, roughly $324\times$ the cost of embedding-based evaluation.
MTG evaluation (100 samples, 6 pairwise comparisons each) adds a further ${\sim}2$--$3$ hours.
This cost disparity highlights a practical advantage of learned identity embeddings: NearID achieves superior discrimination and oracle alignment at a fraction of the inference budget.

\section{Additional Quantitative Results}
\label{sec:supp_quantitative}

\subsection{Loss Decomposition}

\Cref{tab:loss_decomposition} decomposes the training objectives compared in the ablation study (\mainref{tab:ablation}{Table~3 in the main paper}).
All variants share the same frozen backbone and MAP head; only the loss function differs.
The key distinction lies in how NearID distractors ($\mathcal{R}$) enter each formulation:
the standard InfoNCE baseline (\mainref{tab:ablation}{Table~3}, row~1) receives distractors from the data pipeline but \emph{ignores} them entirely in the loss computation~($\varnothing$), relying solely on in-batch negatives.
All other objectives actively use $\mathcal{R}$, either as additional candidates in the softmax denominator~(\textbf{D}), as targets of a separate pairwise or hinge constraint~(\textbf{S}), or exclusively for oracle-supervised ranking~(\textbf{R}).

\begin{table*}[t]
\centering
\caption{
\textbf{Training objective decomposition} (\mainref{tab:ablation}{Table~3 in the main paper}).
Each objective combines a discrimination term $\mathcal{L}_{\text{disc}}$ that separates positives from all negatives, and an optional hierarchy term $\mathcal{L}_{\text{rank}}$ that enforces ordering among negative types, yielding $\mathcal{L} {=} \mathcal{L}_{\text{disc}} {+} \alpha\,\mathcal{L}_{\text{rank}}$.
\textbf{Notation} (cf.\ \mainref{sec:problem_formulation}{Sec.~3.1}):
$\ell_{u,v} {=} \cos(\mathbf{u},\mathbf{v})/\tau$;
$\ell_p$, $\ell_r$, $\ell_b$ denote logits to a positive, a NearID distractor, and a batch negative, respectively;
$\text{LSE}_b {=} \log\!\sum_{g \in \mathcal{B}} e^{\ell_{a,g}}$ (log-sum-exp over batch negatives);
$[\cdot]_+ {=} \max(\cdot, 0)$; $\text{sp}(\cdot) {=} \log(1{+}e^{(\cdot)})$.
\textbf{$\mathcal{R}$ usage} indicates how the NearID distractor set $\mathcal{R}$ enters each loss:
\textbf{D}\,=\,in softmax denominator (competes directly with positives);
\textbf{S}\,=\,separate pairwise or hinge term;
\textbf{R}\,=\,pairwise ranking only;
\textbf{-{}-{}-}\,=\,batch-implicit negatives only (no explicit $\mathcal{R}$ mechanism).
$^\dagger$Representation collapse (\mainref{tab:ablation}{Table~3}).
\textbf{Margin}\,=\,the loss includes an explicit numeric margin $m$ between similarity scores.
\textbf{Oracle}\,=\,the ranking term is supervised by ground-truth edit severity (the ratio of edited area to object area); when \xmark, ranking is unsupervised and purely structural (distractor ${>}$ batch negative), requiring no per-sample annotations.
}
\label{tab:loss_decomposition}
\renewcommand{\arraystretch}{1.25}
\setlength{\tabcolsep}{0.5pt}
\footnotesize
\begin{tabular}{@{} l >{\centering\arraybackslash\scriptsize}p{4.0cm} >{\centering\arraybackslash\scriptsize}p{3.0cm} c c c @{}}
\toprule
\textbf{Training Loss}
  & $\mathcal{L}_{\text{disc}}$
  & $\mathcal{L}_{\text{rank}}$
  & \rotatebox{70}{\textbf{$\mathcal{R}$ usage}}
  & \rotatebox{70}{\textbf{Margin}}
  & \rotatebox{70}{\textbf{Oracle}} \\
\midrule
% ---- (a) Standard contrastive (no hierarchy) ----
InfoNCE (sym.)\,\tabcite{radford2021learning_clip}
  & Symmetric softmax CE over global pool~$\mathcal{G}$; single positive per anchor
  & ---
  & ---
  & \xmark
  & \xmark \\

$+\,\mathcal{R}$\,neg
  & (same); distractors appended as extra negatives in softmax over~$\mathcal{G}$
  & ---
  & D
  & \xmark
  & \xmark \\
$+$\,Oracle$^\dagger$\,\tabcite{burges2005learning_to_rank}
  & (same as InfoNCE)
  & RankNet on oracle-ordered distractor pairs: $-\!\log\sigma(\ell_{r_i} {-} \ell_{r_j} {-} m)$
  & R
  & \cmark
  & \cmark \\

$+\,\mathcal{R}$\,neg $+$\,Oracle
  & (same as $+\,\mathcal{R}$\,neg)
  & (same as $+$\,Oracle)
  & D{+}R
  & \cmark
  & \cmark \\
\addlinespace[3pt]
% ---- (b) Extended losses (distractors in separate constraint) ----
SigLIP$+$Rank\,\tabcite{zhai2023SigLIP}
  & Pairwise sigmoid BCE over~$\mathcal{G}$; $\text{BCE}(\ell_r, 0)$ treats distractors as negatives
  & Log-sigmoid: $-\!\log\sigma(\ell_{r_k} {-} \text{LSE}_b)$
  & S
  & \xmark
  & \xmark \\
Circle$+$Rank$^\dagger$\,\tabcite{sun2020circle_loss}
  & Circle Loss with adaptive weighting; $\mathcal{R}$ pooled with batch negatives
  & Hinge: $\tfrac{1}{K}\sum_k[\ell_b^{\max}{-}\ell_{r_k}{+}m]_+$
  & S
  & \cmark
  & \xmark \\
\addlinespace[3pt]
% ---- (c) NearID loss (ours): distractors in softmax denominator ----
\textbf{$\mathcal{L}_{\textbf{NearID}}$ (Ours)}\tabcite{oord2018representation_infonce_loss}
  & Softmax CE over $\mathcal{G} {\cup} \mathcal{R}$: multi-positive, per-index $-\ell_p {+} \log\!\big(\sum_\mathcal{G} e^{\ell_g} {+} \sum_k e^{\ell_{r_k}}\big)$; (\mainref{eq:disc_avg}{Eq.\,1})
  & Softplus: $\text{sp}(\text{LSE}_b {-} \ell_{r_k})$ (\mainref{eq:rank_avg}{Eq.\,2})
  & D{+}R
  & \xmark
  & \xmark \\
$+$\,Pos.\ Cohesion
  & $+\;\beta\!\cdot\!\tfrac{1}{P}\!\sum_p\!(1 {-} \cos(\mathbf{g}_p, \bar{\mathbf{g}}))$; pulls positives toward prototype~$\bar{\mathbf{g}}$
  & (same)
  & D{+}R
  & \xmark
  & \xmark \\
\bottomrule
\end{tabular}
\end{table*}

\subsection{Correlation Aggregation}
All per-sample Pearson correlation coefficients ($r_k$) are aggregated using Fisher's $z$-transformation~\cite{fisher1915frequency_zfisher} to ensure robustness under varying sample sizes:
\begin{equation}
  \label{eq:fisher_mean_correlation}
  \bar{r} = \tanh\!\left(\frac{1}{N}\sum_{k=1}^{N} \tanh^{-1}(r_k)\right).
\end{equation}
This applies to all reported correlation metrics: \MO, \MOpair, and \MH.

\subsection{Score Distribution Analysis}
\begin{figure}[t]
    \centering
    \begin{subfigure}[t]{0.49\linewidth}
        \centering
        \includegraphics[width=\linewidth]{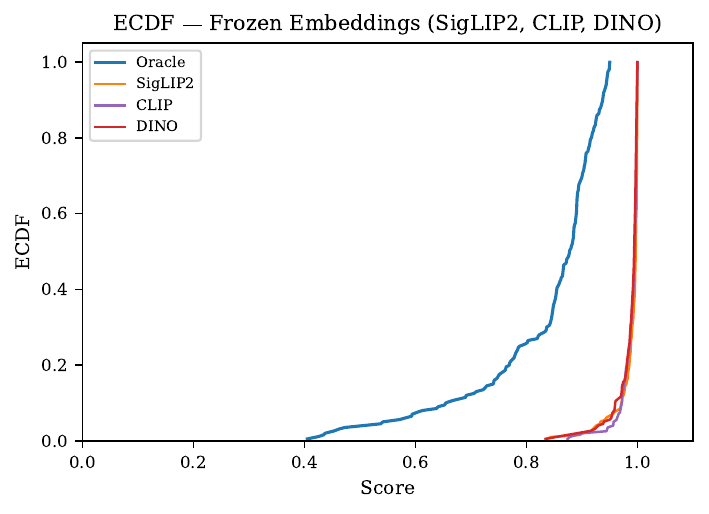}
        \caption{Frozen embedding baselines}
        \label{fig:ecdf_frozen}
    \end{subfigure}\hfill
    \begin{subfigure}[t]{0.49\linewidth}
        \centering
        \includegraphics[width=\linewidth]{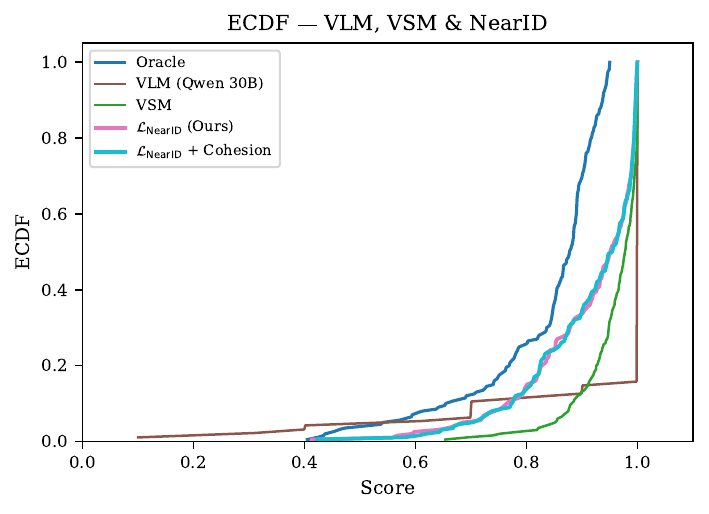}
        \caption{VLM, VSM \& NearID (Ours)}
        \label{fig:ecdf_comparison}
    \end{subfigure}

    \caption{
    \textbf{MTG score distribution alignment with the Oracle.}
    Empirical CDF of per-edit identity-preservation scores on the MTG benchmark~\cite{eldesokeymind_vsm_mtg}.
    The Oracle score is defined as $1 - (|\Omega_{p}| / |\Omega_{o}|)$, where $\Omega_{p}$ and $\Omega_{o}$ are the part and object masks respectively, reflecting the true magnitude of each edit.
    \textbf{(a)}~Frozen embeddings (SigLIP2, CLIP, DINO) cluster near $1.0$ regardless of edit severity, indicating insensitivity to part-level changes.
    \textbf{(b)}~Both NearID configurations produce score distributions that track the Oracle far more closely than VLM or VSM baselines, demonstrating improved sensitivity to fine-grained identity-altering edits.
    }
    \label{fig:mtg_ecdf}
\end{figure}

In \Cref{fig:mtg_ecdf}, we showcase that NearID follows the oracle curve of scores the closest. 
% Included by supp.tex (§Additional Quantitative Results)

\subsection{Per-Source Discrimination Breakdown}

The main results (\mainref{tab:main_table}{Table~1}) report SSR and PA averaged over seven fill-based inpainting sources: four used during training (FLUX.1-Fill and Qwen-Image, each at $512$ and native resolution) and three unseen generators (PowerPaint and SDXL-Inpaint at two resolutions).
\Cref{tab:per_source} provides a complete per-source breakdown, additionally including two Canny-guided inpainting sources (FLUX.1-Canny-Inpaint at $512$ and $1024$) that combine edge-map conditioning with masked fill and are excluded from the main evaluation due to the additional structural prior.
Frozen baselines (SigLIP2, Qwen3-VL~30B, VSM~\cite{eldesokeymind_vsm_mtg}) exhibit large cross-source variance (SSR $12$--$67\%$), whereas NearID maintains ${\geq}\,97.8\%$ SSR across all nine configurations with only a $0.1\%$ gap between training and unseen fill-based averages.

\begin{table*}[t]
\centering
\caption{
\textbf{Per-source discrimination breakdown} (full image).
SSR\,(\%) and PA\,(\%) for each inpainting source used to generate NearID distractors.
The main paper (\mainref{tab:main_results}{Table~1}) reports results averaged over seven fill-based sources; here we additionally include two \emph{Canny-guided} sources (FLUX.1-Canny) that are excluded from the main evaluation.
Training sources ($\dagger$) supply distractors during NearID training; the remaining sources are \emph{unseen}, testing cross-generator generalization.
NearID achieves ${\geq}\,97.8\%$ SSR across all nine configurations, with only $0.1\%$ difference between training and unseen fill-based averages.
}
\label{tab:per_source}
\renewcommand{\arraystretch}{1.15}
\setlength{\tabcolsep}{3pt}
\scriptsize
\begin{tabular}{@{} l l c cc cc cc !{\vrule width 0.5pt} cc @{}}
\toprule
& & & \multicolumn{2}{c}{\textbf{SigLIP2}} & \multicolumn{2}{c}{\textbf{Qwen3-VL}} & \multicolumn{2}{c}{\textbf{VSM}} & \multicolumn{2}{c}{\textbf{NearID (Ours)}} \\
\cmidrule(lr){4-5} \cmidrule(lr){6-7} \cmidrule(lr){8-9} \cmidrule(l){10-11}
\textbf{Generator} & \textbf{Res.} & \textbf{$n$}
  & SSR & PA & SSR & PA & SSR & PA & SSR & PA \\
\midrule
% ---- Training sources ----
\multicolumn{11}{@{}l}{\textit{Training sources}} \\
\addlinespace[2pt]
FLUX.1-Fill$^\dagger$     & 512 & 500 & 35.8 & 55.7 & 48.3 & 68.3 & 23.7 & 38.7 & \textbf{99.4} & \textbf{99.8} \\
FLUX.1-Fill$^\dagger$     & 1024 & 500 & 27.4 & 44.8 & 38.9 & 60.0 & 15.7 & 25.6 & \textbf{99.4} & \textbf{99.7} \\
Qwen-Image$^\dagger$      & 512 & 500 & 33.8 & 51.4 & 48.1 & 67.7 & 27.9 & 42.2 & \textbf{99.2} & \textbf{99.7} \\
Qwen-Image$^\dagger$      & 1328 & 500 & 16.2 & 32.1 & 43.3 & 64.7 & 26.7 & 39.3 & \textbf{98.2} & \textbf{99.5} \\
\addlinespace[4pt]
% ---- Unseen fill-based sources (included in main paper) ----
\multicolumn{11}{@{}l}{\textit{Unseen fill-based generators (included in main results)}} \\
\addlinespace[2pt]
PowerPaint                & 512 & 500 & 34.0 & 52.7 & 67.1 & 83.2 & 51.2 & 66.8 & \textbf{99.2} & \textbf{99.8} \\
SDXL-Inpaint              & 512 & 500 & 42.0 & 59.3 & 60.5 & 79.7 & 54.4 & 72.7 & \textbf{99.8} & \textbf{99.9} \\
SDXL-Inpaint              & 1024 & 500 & 26.0 & 45.6 & 41.9 & 60.7 & 25.3 & 41.5 & \textbf{99.0} & \textbf{99.6} \\
\addlinespace[4pt]
% ---- Unseen Canny-guided sources (excluded from main paper) ----
\multicolumn{11}{@{}l}{\textit{Unseen Canny-guided generators (excluded from main results)}} \\
\addlinespace[2pt]
FLUX.1-Canny              & 512 & 500 & 27.4 & 47.4 & 56.3 & 75.8 & 26.9 & 41.4 & \textbf{97.8} & \textbf{99.1} \\
FLUX.1-Canny              & 1024 & 500 & 25.8 & 47.8 & 40.5 & 62.6 & 12.0 & 22.0 & \textbf{98.8} & \textbf{99.6} \\
\addlinespace[4pt]
\midrule
% ---- Pooled weighted averages ----
Avg.\ (training, 4)       & ---  & 2\,000 & 28.3 & 46.0 & 44.6 & 65.2 & 23.5 & 36.5 & \textbf{99.0} & \textbf{99.7} \\
Avg.\ (unseen fill, 3)    & ---  & 1\,500 & 34.0 & 52.5 & 56.5 & 74.5 & 43.6 & 60.3 & \textbf{99.3} & \textbf{99.8} \\
Avg.\ (unseen canny, 2)   & ---  & 1\,000 & 26.6 & 47.6 & 48.4 & 69.2 & 19.5 & 31.7 & \textbf{98.3} & \textbf{99.3} \\
\addlinespace[2pt]
Avg.\ (all 7, main)       & ---  & 3\,500 & 30.7 & 48.8 & 49.7 & 69.2 & 32.1 & 46.7 & \textbf{99.2} & \textbf{99.7} \\
Avg.\ (all 9)             & ---  & 4\,500 & 29.8 & 48.5 & 49.4 & 69.2 & 29.3 & 43.4 & \textbf{99.0} & \textbf{99.6} \\
\bottomrule
\end{tabular}
\end{table*}

\subsection{Training Data Ablation}
\label{sec:supp_data_ablation}

\Cref{tab:data_ablation} isolates the contribution of each training data component under a fixed loss ($\mathcal{L}_{\text{NearID}}$, \mainref{tab:ablation}{Table~3}).
Three findings emerge.
\textbf{(1)~Near-identity distractors are the dominant signal:} removing the NearID dataset while retaining MTG reduces SSR by $40.8\%$ (from $99.2\%$ to $58.3\%$), whereas removing MTG costs only $0.8\%$.
This confirms that exposure to matched-context distractors during training is essential for learning background-invariant identity representations.
\textbf{(2)~Source diversity matters:} training with a single inpainting engine (FLUX.1-Fill, $1$\,source) versus four engines drops SSR by $11.5\%$ ($98.4\% \to 86.9\%$).
Diverse distractor styles prevent the model from overfitting to generator-specific artifacts.
\textbf{(3)~MTG provides complementary but secondary benefit:} the modest $+0.8\%$ NearID SSR gain from adding MTG suggests that part-level edits refine the embedding space without fundamentally altering its structure, consistent with MTG's role as a fine-grained auxiliary signal.
Conversely, training on MTG data alone yields the highest MTG SSR ($49.0\%$ vs $35.0\%$ for the full pipeline), but at the cost of severely degraded object-level NearID discrimination ($58.3\%$ SSR).
To ensure a fair step-count comparison, the MTG-only setup applies $8\times$ upsampling of the MTG split ($5{,}000$ identities $\times 8 = 40{,}000$ effective samples), which matches the scale of the full joint pipeline (NearID ${\approx}\,20{,}000$ samples $+$ MTG ${\times}\,4 = 40{,}000$ total).
This reveals a critical failure mode: part-level edits alone cannot teach the model to separate different instances sharing the same background, since MTG pairs always depict the \emph{same} object with localized modifications rather than a \emph{different} object in matched context.
The full pipeline strikes the best balance: strong NearID discrimination while retaining meaningful MTG sensitivity ($35.0\%$ SSR, $46.5\%$ PA), confirming that near-identity distractors and part-level edits serve complementary but asymmetric roles, and that the part-level dataset alone is insufficient.

\begin{table}[t]
\centering
\caption{
\textbf{Training data ablation.}
All rows use $\mathcal{L}_{\text{NearID}}$ (\mainref{tab:ablation}{Table~3}) with identical hyperparameters and a frozen SigLIP2 backbone with trainable MAP head. Data scale is accounted for; in the MTG setup, we use x8 to match the number of steps in our main setting. 
\textbf{NearID src} = number of inpainting sources providing near-identity distractors during training;
\textbf{MTG} = whether the MTG part-level dataset is included.
$^\dagger$Object-level discrimination failure: despite the highest MTG SSR/PA, NearID SSR drops to $58.3\%$, indicating that part-level data alone cannot teach background-invariant identity separation.
}
\label{tab:data_ablation}
\renewcommand{\arraystretch}{1.15}
\setlength{\tabcolsep}{2.5pt}
\scriptsize
\begin{tabular}{lcccccccc}
\toprule
\multirow{2}{*}{\textbf{Training Data}}
  & \multirow{2}{*}{\textbf{NearID src}}
  & \multirow{2}{*}{\textbf{MTG}}
  & \multicolumn{2}{c}{\textbf{NearID}}
  & \multicolumn{4}{c}{\textbf{MTG}} \\
\cmidrule(lr){4-5}\cmidrule(lr){6-9}
  & &
  & \textbf{SSR} $\uparrow$
  & \textbf{PA} $\uparrow$
  & \textbf{\MO} $\uparrow$
  & \textbf{\MOpair} $\uparrow$
  & \textbf{SSR} $\uparrow$
  & \textbf{PA} $\uparrow$ \\
\midrule
% ---- Frozen baseline ----
None (frozen)
  & ---
  & ---
  & 30.74
  & 48.81
  & 0.180
  & 0.366
  & 0.0
  & 0.0 \\
\midrule
% ---- Full pipeline (reference row) ----
\textbf{NearID (Ours)}
  & 4
  & \cmark
  & \textbf{99.17}
  & \textbf{99.71}
  & \textbf{0.465}
  & \textbf{0.486}
  & \textbf{35.0}
  & \textbf{46.5} \\
\addlinespace[1pt]
% ---- Deltas relative to full pipeline (inline) ----
% ---- Ablate MTG ----
NearID only
  & 4
  & \xmark
  & 98.37
  & 99.47
  & 0.318
  & 0.457
  & 4.0
  & 6.0 \\
% ---- Ablate NearID ----
MTG only$^\dagger$
  & 0
  & \cmark
  & 58.34
  & 75.64
  & 0.349
  & 0.395
  & \textit{49.0}
  & \textit{62.0} \\
% ---- Ablate source diversity ----
NearID$_{1}$ only
  & 1
  & \xmark
  & 86.86
  & 93.77
  & 0.326
  & 0.456
  & 3.0
  & 5.5 \\
\bottomrule
\end{tabular}
\end{table}

\subsection{Role-Aware Foreground Masking}
\label{sec:supp_role_masking}
During training we black out background pixels (outside the oracle foreground mask) at role-specific rates: anchors $p{=}0.5$, positives $p{=}0.2$, distractors $p{=}0.6$.
Positives vary background by design, so heavy masking would discard the cross-background context that defines a positive pair; distractors share the anchor's \emph{exact} background, so elevated masking prevents inpainting-boundary artifacts from becoming trivial rejection cues.
To verify masking is necessary, we retrain a mask-free variant ($p{=}0$, otherwise identical); \Cref{tab:fg_masking} reports the comparison.
Without masking, foreground-only oracle alignment drops (\MO\ $0.391$ vs.\ $0.641$) and MTG SSR roughly halves under foreground evaluation ($21\%$ vs.\ $32\%$), revealing reliance on background context; with role-aware masking the model instead improves \MO\ from full to foreground evaluation ($0.465 \to 0.641$) and holds MTG SSR ($35 \to 32\%$), confirming it learns foreground identity rather than background shortcuts.
Both trained variants reach ${\sim}99\%$ full-image NearID SSR, so masking is not what drives near-identity discrimination; its effect is visible only once the background is removed.

\begin{table}[t]
\centering
\caption{
\textbf{Role-aware foreground masking ablation.}
Both trained rows use $\mathcal{L}_{\text{NearID}}$ (\mainref{tab:ablation}{Table~3}) with identical hyperparameters and a frozen SigLIP2 backbone with trainable MAP head; only the role-specific masking probabilities differ.
\textbf{Full} = standard evaluation on the complete image;
\textbf{FG} = foreground-only evaluation with the background removed via the oracle segmentation mask.
\textbf{NearID SSR} = near-identity discrimination;
\textbf{\MO} = Pearson correlation with the MTG part-level oracle;
\textbf{MTG SSR} = part-level discrimination.
Both trained variants reach ${\sim}99\%$ full-image NearID SSR, so the diagnostic signal lies in the FG columns: without masking \MO\ drops and MTG SSR roughly halves, revealing reliance on background context.
}
\label{tab:fg_masking}
\renewcommand{\arraystretch}{1.15}
\setlength{\tabcolsep}{2.5pt}
\scriptsize
\begin{tabular}{lcccccc}
\toprule
\multirow{2}{*}{\textbf{Method}}
  & \multicolumn{2}{c}{\textbf{NearID SSR} $\uparrow$}
  & \multicolumn{2}{c}{\textbf{\MO} $\uparrow$}
  & \multicolumn{2}{c}{\textbf{MTG SSR} $\uparrow$} \\
\cmidrule(lr){2-3}\cmidrule(lr){4-5}\cmidrule(lr){6-7}
  & \textbf{Full} & \textbf{FG}
  & \textbf{Full} & \textbf{FG}
  & \textbf{Full} & \textbf{FG} \\
\midrule
% ---- Frozen baseline ----
SigLIP2 (frozen)
  & 30.7 & 64.7
  & 0.179 & 0.365
  & 0.0 & 6.0 \\
\midrule
% ---- Full pipeline (reference row) ----
\textbf{NearID (Ours)}
  & 99.2 & \textbf{95.3}
  & 0.465 & \textcolor{cmark-green}{\textbf{0.641}}
  & 35.0 & \textbf{32.0} \\
% ---- Mask-free variant (p=0 for all roles) ----
~~($-$~masking)
  & 99.8 & 95.4
  & 0.512 & \textcolor{xmark-red}{\textbf{0.391}}
  & 43.0 & \textcolor{xmark-red}{\textbf{21.0}} \\
\bottomrule
\end{tabular}
\end{table}

\subsection{Positive Cohesion Variant}
\label{sec:supp_cohesion}

The ``$+$\,Pos.\ Cohesion'' variant in \mainref{tab:ablation}{Table~3} extends $\mathcal{L}_{\text{NearID}}$ (\mainref{eq:nearid_loss}{Eq.\,3}) with a term that explicitly pulls multiple positive views of the same identity toward a shared prototype.
Given $P$ valid positive embeddings $\{\mathbf{g}_{i,p}\}_{p=1}^{P}$ for anchor $i$, we compute the $\ell_2$-normalized prototype
\begin{equation}
  \bar{\mathbf{g}}_i = \frac{\sum_{p=1}^{P} \mathbf{g}_{i,p}}{\big\|\sum_{p=1}^{P} \mathbf{g}_{i,p}\big\|_2},
  \label{eq:prototype}
\end{equation}
and define the cohesion loss as the average cosine distance of each positive to this prototype:
\begin{equation}
  \mathcal{L}_{\text{coh}} = \frac{1}{P}\sum_{p=1}^{P}\bigl(1 - \cos(\mathbf{g}_{i,p},\, \bar{\mathbf{g}}_i)\bigr).
  \label{eq:cohesion}
\end{equation}
The total objective becomes
$\mathcal{L} = \mathcal{L}_{\text{NearID}} + \beta\,\mathcal{L}_{\text{coh}}$
with $\beta = 0.1$ (see \Cref{sec:supp_beta_ablation} for an ablation over $\beta$).
Samples with fewer than two valid positives contribute zero cohesion loss.

The rationale is that $\mathcal{L}_{\text{disc}}$ (\mainref{eq:disc_avg}{Eq.\,1}) separates each positive individually from negatives but does not explicitly reduce variance \emph{among} positives.
$\mathcal{L}_{\text{coh}}$ was hypothesised to directly tighten the positive cluster, improving intra-identity compactness and reducing cross-identity overlap in the embedding space.

%SUBMITTED VERSION: In practice (\mainref{tab:ablation}{Table~3}), the cohesion variant yields slightly higher NearID SSR ($99.3\%$ vs $99.2\%$) and the best MTG SSR ($51.0\%$ vs $35.0\%$), suggesting improved part-level sensitivity.
%SUBMITTED VERSION: However, NearID PA, \MO, and \MH\ remain comparable, indicating that the base $\mathcal{L}_{\text{NearID}}$ already produces sufficiently compact positive clusters for the primary identity discrimination task.
% FIX: R9 MTG SSR was 51.0 (SSR-mean) — corrected to 36.0 (SSRm AND-based) to match all other rows.
In practice (\mainref{tab:ablation}{Table~3}), the cohesion variant yields slightly higher NearID SSR ($99.3\%$ vs $99.2\%$) and comparable MTG SSR ($36.0\%$ vs $35.0\%$).
However, NearID PA, \MO, and \MH\ remain comparable, indicating that the base $\mathcal{L}_{\text{NearID}}$ already produces sufficiently compact positive clusters for the primary identity discrimination task.

\subsection{Extended Baselines: Foreground Masking and VLM Scaling}
\label{sec:supp_extended_baselines}

To disentangle background dependence from identity-discrimination capacity, we evaluate all methods under two conditions: \textbf{Full}~image (standard) and \textbf{FG}~only (foreground extracted via oracle segmentation mask, background removed).
We additionally report Qwen3-VL at three model scales (4B, 8B, 30B) to assess whether VLM scaling can compensate for background confusion.
Beyond object-level discrimination (SSR/PA), we also report the MTG oracle correlation (\MO) under both conditions for embedding-based methods, measuring how well each model's similarity scores track the true magnitude of part-level edits when background context is removed.

\Cref{tab:extended_baselines} reveals four key findings.
\textbf{(1)~Frozen embeddings are heavily background-dependent:} removing the background improves CLIP by $+41.0\%$ SSR, DINOv2 by $+43.3\%$, and SigLIP2 by $+33.9\%$, confirming that these encoders rely substantially on contextual cues when the foreground instance is ambiguous.
\textbf{(2)~VLMs benefit from masking but with diminishing returns at scale:} Qwen3-VL\,4B gains $+11.0\%$, 8B gains $+13.5\%$, but 30B gains only $+5.5\%$ SSR from background removal.
This suggests that larger VLMs partially learn to discount background cues through their language-grounded reasoning, but remain susceptible to matched-context confusion; even explicit anti-background prompting (\Cref{sec:supp_vlm_prompt}) does not fully resolve the problem.
The MTG oracle alignment (\MO) under FG masking shows an inconsistent pattern across VLM scales ($4$B: $-0.05$; $8$B: $+0.12$; $30$B: $-0.05$), suggesting that the ability to leverage part-level similarity cues does not improve monotonically with model scale or background removal.
\textbf{(3)~NearID is inherently background-invariant for discrimination:} NearID shows a slight SSR \emph{decrease} ($-3.8\%$) under foreground masking, indicating that it has learned to extract identity features from full images without background shortcuts.
This is qualitatively distinct from the frozen encoders, which rely so heavily on contextual cues that removing the background yields gains of $+34$--$43\%$ SSR: for NearID the background is not a shortcut but a complementary signal that the model incorporates without becoming dependent on it, so removing it causes a minor performance regression from mask-boundary information loss rather than from background dependence.
VSM~\cite{eldesokeymind_vsm_mtg} is similarly unaffected, consistent with its internal use of foreground segmentation; its \MO\ ($0.394$) is unchanged under external masking.
\textbf{(4)~Background removal improves part-level oracle alignment:} despite stable discrimination, NearID's \MO\ correlation with MTG oracle scores improves substantially under FG masking ($0.465 \to 0.641$, $+0.18$).
This indicates that while NearID already discriminates objects correctly regardless of background, the background context still introduces noise into the \emph{magnitude} of similarity differences for part-level edits.
Frozen SigLIP2 shows a similar \MO\ gain ($+0.19$), while DINOv2 is unaffected ($-0.02$).
The cohesion variant shows the highest FG \MO\ ($0.648$, vs.\ $0.641$ for NearID), a marginal gain that mirrors its small MTG SSR improvement ($36.0\%$ vs.\ $35.0\%$) observed in the ablation study (\mainref{tab:ablation}{Table~3}).

\begin{table}[t]
\centering
\caption{
\textbf{Effect of foreground masking and VLM scaling} on NearID discrimination and MTG oracle alignment.
\textbf{Full}: evaluation on unmodified images.
\textbf{FG}: evaluation on foreground-only crops (background removed via oracle mask).
$\Delta$: absolute change from masking (FG\,$-$\,Full).
Frozen embeddings gain $+34$--$43\%$ SSR from background removal, confirming strong background dependence.
VLMs also benefit ($+5$ to $+14\%$), with diminishing gains at larger scale.
NearID SSR is nearly unaffected ($-3.8\%$), validating inherent background invariance.
Notably, NearID oracle alignment (\MO) improves substantially under FG masking ($+0.18$), suggesting that background context adds noise to part-level correlation even when object-level discrimination is already robust.
$^\dagger$VSM uses foreground masks internally, so external masking has no additional effect.
$^\ddagger$Qwen3-VL \MH\ on DB++ is omitted from \mainref{tab:main_table}{Table~1} as the autoregressive VLM evaluation on DB++ was skipped due to computational cost (${\sim}324\times$ embedding-based evaluation).
}
\label{tab:extended_baselines}
\renewcommand{\arraystretch}{1.15}
\setlength{\tabcolsep}{2.5pt}
\scriptsize
\begin{tabular}{l ccc ccc cc}
\toprule
\multirow{2}{*}{\textbf{Method}} &
\multicolumn{3}{c}{\textbf{Full Image}} &
\multicolumn{3}{c}{\textbf{FG Only}} &
\multicolumn{1}{c}{$\boldsymbol{\Delta}$} &
\multicolumn{1}{c}{$\boldsymbol{\Delta}$} \\
\cmidrule(lr){2-4}\cmidrule(lr){5-7}
& \textbf{SSR}$\uparrow$ & \textbf{PA}$\uparrow$ & \textbf{\MO}$\uparrow$
& \textbf{SSR}$\uparrow$ & \textbf{PA}$\uparrow$ & \textbf{\MO}$\uparrow$
& \textbf{SSR} & \textbf{\MO} \\
\midrule
% ---- Frozen embedding baselines ----
CLIP ViT-L/14
  & 10.3 & 20.9 & 0.239
  & 51.3 & 71.6 & 0.297
  & \posdelta{+41.0} & \posdelta{+0.06} \\
DINOv2 ViT-L/14
  & 20.4 & 34.6 & 0.324
  & 63.7 & 77.4 & 0.301
  & \posdelta{+43.3} & \negdelta{$-$0.02} \\
SigLIP2-SO400M
  & 30.7 & 48.8 & 0.179
  & 64.7 & 80.5 & 0.365
  & \posdelta{+33.9} & \posdelta{+0.19} \\
VSM~\cite{eldesokeymind_vsm_mtg}
  & 32.1 & 46.7 & 0.394
  & 32.0 & 46.7 & 0.394$^\dagger$
  & \negdelta{$-$0.1} & $\approx$0 \\
\midrule
% ---- VLM baselines (Qwen3-VL scaling) ----
Qwen3-VL\,4B~\cite{bai2025qwen3}
  & 13.4 & 32.9 & 0.012
  & 24.4 & 44.3 & $-$0.046
  & \posdelta{+11.0} & \negdelta{$-$0.06} \\
Qwen3-VL\,8B
  & 27.5 & 46.9 & 0.134
  & 40.9 & 58.1 & 0.252
  & \posdelta{+13.5} & \posdelta{+0.12} \\
Qwen3-VL\,30B
  & 49.7 & 69.2 & 0.219
  & 55.2 & 74.2 & 0.173
  & \posdelta{+5.5} & \negdelta{$-$0.05} \\
\midrule
% ---- NearID (ours) ----
\textbf{NearID (Ours)}
  & 99.2 & \textbf{99.7} & \textbf{0.465}
  & \textbf{95.3} & \textbf{98.0} & \textbf{0.641}
  & \negdelta{$-$3.8} & \posdelta{+0.18} \\
% ~~+ Pos.\ Cohesion
%   & \textbf{99.3} & \textbf{99.7} & 0.459
%   & 95.3 & \textbf{98.0} & \textbf{0.648}
%   & \negdelta{$-$4.0} & \posdelta{+0.19} \\
\bottomrule
\end{tabular}
\end{table}

\subsection{Ranking Weight Ablation ($\alpha$)}
\label{sec:supp_alpha_ablation}

The NearID loss combines a discrimination term $\mathcal{L}_{\text{disc}}$ with a ranking regulariser $\mathcal{L}_{\text{rank}}$ weighted by $\alpha$ (\mainref{eq:nearid_loss}{Eq.\,3}).
\Cref{tab:alpha_ablation} varies $\alpha$ while holding all other hyperparameters fixed.
Even without the ranking term ($\alpha{=}0$), the discrimination loss alone achieves near-perfect NearID SSR ($99.6\%$) because near-identity distractors already participate in the softmax denominator.
Oracle alignment (\MO) generally increases with $\alpha$ up to $1.0$, while discrimination (SSR) degrades mildly, revealing a clear trade-off between graded similarity structure and discrimination sharpness.
At $\alpha{=}2.0$ both SSR and \MO decline, indicating that an overly dominant ranking signal begins to collapse the representation.
We adopt $\alpha{=}0.5$ as the default configuration in the main paper (\mainref{tab:main_table}{Table~1}), as it provides the most balanced operating point: near-perfect discrimination (SSR\,$=$\,$99.2\%$, within $0.4\%$ of the $\alpha{=}0$ ceiling) while achieving an \MO\ of $0.465$, more than $1.5{\times}$ the alignment obtained without ranking ($\alpha{=}0$: $0.306$).
Among the values studied, $\alpha{=}1.0$ achieves the highest oracle alignment ($\MO{=}0.542$) and second-best human alignment ($\MH{=}0.571$, narrowly behind $\alpha{=}2.0$ at $0.573$ but with substantially better SSR and \MO), making it the operating point with the strongest stable alignment signal.
We therefore study how positive cohesion ($\beta$) interacts with the ranking term at this value to provide guidance on hyperparameter choices in the high-alignment regime.

\begin{table}[t]
\centering
\caption{
\textbf{Effect of the ranking weight $\alpha$} in the NearID loss
$\mathcal{L}_{\text{NearID}} = \mathcal{L}_{\text{disc}} + \alpha\,\mathcal{L}_{\text{rank}}$.
All variants use a frozen SigLIP2 backbone with a trainable MAP head, trained on NearID\,+\,MTG data.
$\alpha{=}0$ ablates the ranking term entirely, retaining only the discrimination loss with near-identity distractors in the softmax denominator.
Oracle alignment (\MO) generally increases with $\alpha$ up to $1.0$, while discrimination (SSR) degrades only mildly until $\alpha{=}1.0$ and drops at $\alpha{=}2.0$.
Human-judgment alignment (\MH) rises sharply at $\alpha{=}0.1$ and remains broadly stable ($0.54$--$0.57$) for $\alpha{\geq}0.25$, confirming that the ranking term is essential for perceptually meaningful embeddings.
}
\label{tab:alpha_ablation}
\renewcommand{\arraystretch}{1.15}
\setlength{\tabcolsep}{3.5pt}
\small
\begin{tabular}{lccccccc}
\toprule
\multirow{2}{*}{\textbf{$\alpha$}} &
\multicolumn{2}{c}{\textbf{NearID}} &
\multicolumn{4}{c}{\textbf{MTG}} &
\textbf{DB++} \\
\cmidrule(lr){2-3}\cmidrule(lr){4-7}\cmidrule(lr){8-8}
& \textbf{SSR} $\uparrow$ & \textbf{PA} $\uparrow$ &
\textbf{\MO} $\uparrow$ & \textbf{\MOpair} $\uparrow$ &
\textbf{SSR} $\uparrow$ & \textbf{PA} $\uparrow$ &
\textbf{\MH} $\uparrow$ \\
\midrule
$\alpha{=}0$ (disc.~only)
  & \textbf{99.57} & \textbf{99.85}
  & 0.306 & 0.336 & \textbf{63.0} & \textbf{73.0}
  & 0.229 \\
$\alpha{=}0.1$
  & \textbf{99.57} & 99.84
  & 0.362 & 0.422 & 58.0 & 69.0
  & 0.521 \\
$\alpha{=}0.25$
  & 99.20 & 99.73
  & 0.466 & 0.518 & 49.0 & 62.0
  & 0.555 \\
$\alpha{=}0.5$ (default)
  & 99.17 & 99.71
  & 0.465 & 0.486 & 35.0 & 46.5
  & 0.545 \\
$\alpha{=}1.0$
  & 98.80 & 99.61
  & \textbf{0.542} & \textbf{0.583} & 35.0 & 49.5
  & 0.571 \\
$\alpha{=}2.0$
  & 97.09 & 98.94
  & 0.494 & 0.562 & 26.0 & 34.5
  & \textbf{0.573} \\
\bottomrule
\end{tabular}
\end{table}

\subsection{Positive Cohesion Weight Ablation ($\beta$)}
\label{sec:supp_beta_ablation}

The combined objective $\mathcal{L}_{\text{NearID}} + \beta\,\mathcal{L}_{\text{coh}}$ (\Cref{sec:supp_cohesion}) extends $\mathcal{L}_{\text{NearID}}$ with a positive cohesion term weighted by $\beta$, designed to draw multi-view embeddings of the same identity closer via a prototype-centred objective.
\Cref{tab:beta_ablation} evaluates $\beta \in \{0, 0.1, 0.2, 0.3\}$ at $\alpha{=}1.0$, where the ranking term exerts its strongest stable influence.
Contrary to expectation, adding positive cohesion ($\beta > 0$) consistently degrades oracle alignment ($\MO$), MTG discrimination (SSR), and human alignment ($\MH$) relative to $\beta{=}0$.
NearID SSR remains nearly unchanged (${\approx}98.8\%$), indicating that the cohesion signal does not improve identity discrimination but instead perturbs the representation away from the graded similarity structure learned by the ranking term alone.
We conclude that at $\alpha{=}1.0$ the ranking term already provides sufficient pull towards identity-consistent embeddings; adding an explicit cohesion loss introduces redundant gradient signal that interferes with fine-grained alignment.
This conclusion extends to the default $\alpha{=}0.5$ setting: in \mainref{tab:ablation}{Table~3}, the $+$\,Pos.\,Cohesion variant (R9, $\alpha{=}0.5$, $\beta{=}0.1$) yields $\MO{=}0.459$ versus $0.465$ for the base $\mathcal{L}_{\text{NearID}}$, confirming that the ranking term at $\alpha{=}0.5$ already provides sufficient alignment signal and that $\beta{>}0$ is detrimental across both operating points.
These results further support $\alpha{=}0.5$ (without cohesion) as the recommended default.

\begin{table}[t]
\centering
\caption{
\textbf{Effect of positive cohesion weight $\beta$} in the combined objective
$\mathcal{L}_{\text{NearID}} + \beta\,\mathcal{L}_{\text{coh}}$ (\Cref{sec:supp_cohesion}),
evaluated at $\alpha{=}1.0$.
All variants use a frozen SigLIP2 backbone with a trainable MAP head, trained on NearID\,+\,MTG data.
$\beta{=}0$ (no cohesion) achieves the best oracle alignment (\MO) and human alignment (\MH\,DB++);
increasing $\beta$ degrades all alignment metrics without improving discrimination (SSR).
}
\label{tab:beta_ablation}
\renewcommand{\arraystretch}{1.15}
\setlength{\tabcolsep}{3.5pt}
\small
\begin{tabular}{lccccccc c}
\toprule
\multirow{2}{*}{\textbf{$\beta$ (cohesion)}} &
\multicolumn{2}{c}{\textbf{NearID}} &
\multicolumn{4}{c}{\textbf{MTG}} &
\multicolumn{1}{c}{\textbf{DB++}} \\
\cmidrule(lr){2-3}\cmidrule(lr){4-7}\cmidrule(lr){8-8}
& \textbf{SSR} $\uparrow$ & \textbf{PA} $\uparrow$ &
\textbf{\MO} $\uparrow$ & \textbf{\MOpair} $\uparrow$ &
\textbf{SSR} $\uparrow$ & \textbf{PA} $\uparrow$ &
\textbf{\MH} $\uparrow$ \\
\midrule
$\beta{=}0$ \little{(no cohesion, $\alpha{=}1.0$)}
  & \textbf{98.80} & 99.61
  & \textbf{0.542} & \textbf{0.583} & \textbf{35.0} & \textbf{49.5}
  & \textbf{0.571} \\
$\beta{=}0.1$
  & 98.94 & \textbf{99.67}
  & 0.442 & 0.478 & 30.0 & 40.0
  & 0.555 \\
$\beta{=}0.2$
  & 98.74 & 99.59
  & 0.444 & 0.482 & 29.0 & 37.5
  & 0.550 \\
$\beta{=}0.3$
  & 98.94 & 99.68
  & 0.430 & 0.468 & 29.0 & 36.5
  & 0.553 \\
\bottomrule
\end{tabular}
\end{table}

\subsection{InfoNCE Component Ablation: Distractors and Ranking}
\label{sec:supp_infonce_ablation}

\mainref{tab:ablation}{Table~3} presents four InfoNCE variants that progressively add near-identity distractors ($\mathcal{R}$\,neg) and oracle-supervised ranking to a standard symmetric InfoNCE baseline.
\Cref{tab:infonce_ablation} isolates these four rows alongside the frozen backbone and the full NearID loss, with DB++~\MH\ values now available for all variants (\Cref{tab:loss_decomposition} for mathematical definitions of each component).

\begin{table}[t]
\centering
\caption{
\textbf{InfoNCE component ablation.}
Rows extracted from \mainref{tab:ablation}{Table~3} with DB++~\MH\ now reported for all variants.
``$\mathcal{R}$\,neg'' adds near-identity distractors to the softmax denominator;
``Oracle Ranking'' adds a pairwise RankNet term supervised by oracle similarity.
$\mathcal{L}_{\text{NearID}}$ differs in three ways: symmetric multi-positive softmax, distractors in the denominator, and an \emph{unsupervised} softplus ranking regulariser (see \Cref{tab:loss_decomposition}).
$^\dagger$Representation collapse.
}
\label{tab:infonce_ablation}
\renewcommand{\arraystretch}{1.15}
\setlength{\tabcolsep}{2.5pt}
\small
\begin{tabular}{lcccccc}
\toprule
\multirow{2}{*}{\textbf{Variant}} &
\multicolumn{2}{c}{\textbf{NearID}} &
\multicolumn{2}{c}{\textbf{MTG}} &
\multicolumn{2}{c}{\textbf{DB++}} \\
\cmidrule(lr){2-3}\cmidrule(lr){4-5}\cmidrule(lr){6-7}
& \textbf{SSR} $\uparrow$ & \textbf{PA} $\uparrow$ &
\textbf{\MO} $\uparrow$ & \textbf{\MOpair} $\uparrow$ &
\textbf{\MH} $\uparrow$ & $\Delta$\textbf{\MH} \\
\midrule
None (frozen)
  & 30.74 & 48.81
  & 0.180 & 0.366
  & 0.516 & --- \\
\midrule
InfoNCE \little{(sym., 1-pos)} % R1
  & 60.97 & 75.26
  & 0.267 & 0.418
  & 0.555 & (ref.) \\
~~$+\,\mathcal{R}$\,neg % R1-neg
  & 99.57 & 99.79
  & 0.236 & 0.267
  & 0.251 & $-0.304$ \\
~~$+$\,Oracle Ranking$^\dagger$ % R2
  & 86.34 & 92.25
  & 0.299$^\dagger$ & 0.444$^\dagger$
  & 0.167$^\dagger$ & $-0.388$ \\
~~$+\,\mathcal{R}$\,neg $+$\,Oracle % R2-neg
  & 99.60 & 99.89
  & 0.247 & 0.277
  & 0.227 & $-0.328$ \\
\midrule
\textbf{$\mathcal{L}_{\textbf{NearID}}$ (Ours)} % R7
  & \textbf{99.17} & \textbf{99.71}
  & \textbf{0.465} & \textbf{0.486}
  & \textbf{0.545} & $-0.010$ \\
\bottomrule
\end{tabular}
\end{table}

The component decomposition reveals three distinct mechanisms and their failure modes.

\textbf{(1)~Distractors in the denominator drive discrimination but degrade alignment.}
Adding near-identity distractors as additional negatives in the softmax denominator ($+\,\mathcal{R}$\,neg, R1-neg) raises NearID SSR from $61.0\%$ to $99.6\%$ and PA from $75.3\%$ to $99.8\%$, confirming that the mere presence of confusable negatives in the contrastive pool is sufficient for near-perfect discrimination.
However, this comes at a substantial cost: \MO\ drops from $0.267$ to $0.236$ and \MH\ from $0.555$ to $0.251$.
The model learns a binary identity decision boundary that ignores graded perceptual similarity, producing embeddings poorly aligned with both oracle scores and human judgments.

\textbf{(2)~Oracle ranking alone is insufficient without distractors in the denominator.}
The $+$\,Oracle Ranking variant (R2) adds a RankNet-style pairwise ranking loss over oracle-ordered distractors without placing them in the softmax denominator.
This moderately improves SSR ($61.0\% \to 86.3\%$) but does not close the gap to near-perfect discrimination, and collapses the representation: \MH\ drops to $0.167$ (below even the frozen backbone's $0.516$), indicating severe overfitting to the ranking signal.
Combining both mechanisms ($+\,\mathcal{R}$\,neg $+$\,Oracle, R2-neg) recovers discrimination ($99.6\%$ SSR) but \MH\ remains degraded at $0.227$, showing that oracle-supervised pairwise ranking on top of a one-directional InfoNCE base does not yield well-calibrated embeddings.

\textbf{(3)~The NearID formulation resolves the discrimination--alignment trade-off.}
$\mathcal{L}_{\text{NearID}}$ (\mainref{eq:nearid_loss}{Eq.\,3}) achieves $99.2\%$ SSR while maintaining the highest \MH\ ($0.545$) among all trained variants, and an \MO\ of $0.465$ that is more than $1.5{\times}$ higher than any pure-InfoNCE variant.
The critical difference is architectural: NearID uses a multi-positive symmetric softmax with distractors in the denominator (not one-directional) and replaces pairwise oracle ranking with an unsupervised softplus ranking regulariser $\mathcal{L}_{\text{rank}}$ (\mainref{eq:rank_avg}{Eq.\,2}) that encourages distractors to rank above batch negatives without requiring oracle labels.
This combination preserves graded similarity structure (high \MO, \MH) while achieving comparable discrimination to the brute-force denominator approach.
The $\alpha$-sweep (\Cref{tab:alpha_ablation}) further confirms that the ranking term controls the discrimination--alignment trade-off smoothly, with $\alpha{=}0.5$ providing the best balance for the primary setting.

\subsection{VLM Judge Prompt Template}
\label{sec:supp_vlm_prompt}

Recent work has shown that multimodal LLMs struggle with multi-image comparison tasks, frequently failing to detect fine-grained differences between visually similar image pairs~\cite{ku2024viescore}.
This limitation is particularly relevant for identity evaluation under matched-context conditions, where two images share the same background but differ only in the foreground instance.
VLM behavior can be strongly influenced by background patterns~\cite{vo2025vision}, and when presented with full images, VLMs tend to conflate scene-level similarity with instance-level identity.

To provide the strongest possible VLM baseline, we evaluate Qwen3-VL\,30B~\cite{bai2025qwen3} with a carefully designed structured prompt (\Cref{fig:vlm_prompt}) that explicitly instructs the model to (i)~ignore background and scene cues, (ii)~focus only on object-instance evidence such as unique markings and fine geometry, and (iii)~distinguish instance identity from category-level similarity.
The model receives two images and returns a JSON object containing instance-level match/conflict cues, a confidence flag, and an integer score on a 0--10 rubric.
All Qwen3-VL models (4B, 8B, 30B) are loaded in BF16 precision with FlashAttention-2 for memory-efficient inference.
Because the structured JSON response (with evidence lists and metadata) typically spans ${\sim}250$ tokens, we set the maximum generation length to $512$ tokens; shorter budgets cause truncated output and unparseable responses, silently degrading evaluation coverage.
We map the VLM score to a cosine-similarity proxy via $s = \text{score}/10$ so that the same SSR/PA evaluation protocol applies uniformly across all methods.

Despite this optimized prompting strategy, Qwen3-VL\,30B achieves only $49.7\%$ SSR on our NearID benchmark (\mainref{tab:main_table}{Table~1}), compared to $99.2\%$ for NearID\@.
While the VLM substantially outperforms frozen embedding baselines (SigLIP2: $30.7\%$), the gap to NearID confirms that even state-of-the-art VLMs with explicit anti-background instructions remain susceptible to matched-context confusion, underscoring the need for specialized identity embeddings trained with near-identity distractors.

\begin{figure}[t]
\centering
\tikzset{
  promptbox/.style={
    draw=black!30,
    fill=black!3,
    rounded corners=3pt,
    inner sep=6pt,
    text width=\linewidth-16pt,
    font=\scriptsize\ttfamily,
  }
}
\begin{tikzpicture}
\node[promptbox] (box) {%
\raggedright
You are a rigorous visual evaluator for INSTANCE identity (same physical object), not category.\\[3pt]
You will see TWO images:\\
- Image A\\
- Image B\\[4pt]
\textrm{\textbf{\scriptsize TASK:}}\\[2pt]
Decide whether A and B show the SAME physical object instance (e.g., the same specific car), ignoring background changes and typical viewpoint/lighting changes.\\[4pt]
\textrm{\textbf{\scriptsize IMPORTANT:}}\\[2pt]
- Do NOT use background, scene, or watermark cues as evidence.\\
- Do NOT decide based on category-level similarity (e.g., "both are sedans", "both are red").\\
- Use only object-instance cues visible on the subject itself.\\[4pt]
\textrm{\textbf{\scriptsize RUBRIC (0--10):}}\\[2pt]
10 = definitely same instance (multiple strong, unique matching cues; no meaningful conflicts)\\
7--9 = likely same instance (several matching distinctive cues; minor uncertainty)\\
4--6 = uncertain (insufficient visibility/occlusion/blur; cues are generic)\\
1--3 = likely different instances (distinctive conflicts: geometry, markings, parts, proportions)\\
\phantom{1}0 = definitely different instances (clear contradictory unique traits)\\[4pt]
\textrm{\textbf{\scriptsize EVIDENCE PRIORITY (use these, in order):}}\\[2pt]
(1) Unique markings: scratches, dents, decals, stains, wear patterns, rust, custom mods\\
(2) Fine geometry/parts: headlight shapes, grille pattern, rim design, handle shape, seams\\
(3) Text/logos/serial-like details if clearly readable\\
(4) Color/paint only if it includes unique patterns (not plain "red/blue")\\[4pt]
\textrm{\textbf{\scriptsize HANDLE DIFFICULT CASES:}}\\[2pt]
- If the subject is partially visible, small, blurred, or occluded: score \~{}5 and say "insufficient evidence".\\
- If differences can plausibly be viewpoint/lighting artifacts, do NOT over-penalize; explain uncertainty.\\[4pt]
\textrm{\textbf{\scriptsize OUTPUT:}}\\[2pt]
Return JSON ONLY, exactly matching this schema:\\
\{\\
~~"match\_cues": ["...","..."],\\
~~"conflict\_cues": ["...","..."],\\
~~"background\_used": false,\\
~~"confidence": "high" | "medium" | "low",\\
~~"score": <number from 0 to 10>\\
\}\\[3pt]
\textrm{\textbf{\scriptsize Rules:}}\\[2pt]
- score must be a NUMBER (not a string).\\
- match\_cues/conflict\_cues must refer to the OBJECT, not the background.\\
- background\_used must be false unless you truly cannot avoid it; if true, set confidence="low".\\
- Do not wrap the JSON in markdown fences.\\
- Do not output any text before or after the JSON object.\\
- Concise wording for both match\_cues and conflict\_cues
};
\end{tikzpicture}
\caption{
\textbf{VLM judge prompt template} used for the Qwen3-VL\,30B baseline.
The prompt enforces instance-level (not category-level) identity judgment with an explicit evidence hierarchy and structured JSON output.
}
\label{fig:vlm_prompt}
\end{figure}

\subsection*{Scope and Future Directions}

The NearID benchmark is designed for \emph{concept-preservation} evaluation, where the primary criterion is whether a generated or retrieved image depicts the correct object instance, independently of scene context.
This setting reflects the predominant use case in subject-driven generation~\cite{ruiz2023dreambooth, peng2024dreambench++}, where faithfulness to the reference identity is the central objective.
Text-guided \emph{editing} techniques, however, impose a qualitatively different requirement: the output must simultaneously satisfy the textual editing instruction and preserve the identity of the source concept.
These two objectives are often in tension, and characterising the appropriate trade-off is not addressed by the current benchmark, whose near-identity distractors hold the editing directive fixed by construction.
A rigorous evaluation of this balance would require a dedicated editing benchmark comprising paired source and result images, explicit textual instructions, quality-filtered outputs from state-of-the-art editing methods, and human judgements of both edit-intent adherence and concept fidelity.
Extending the NearID evaluation framework to this setting is a natural direction for future work.

\end{document}